\documentclass[a4paper,fleqn]{cas-sc}

\usepackage[authoryear,longnamesfirst]{natbib}

\def\tsc#1{\csdef{#1}{\textsc{\lowercase{#1}}\xspace}}
\tsc{WGM}
\tsc{QE}
\tsc{EP}
\tsc{PMS}
\tsc{BEC}
\tsc{DE}

\usepackage{cleveref}
\usepackage[acronym]{glossaries}
\usepackage{pifont}
\usepackage{stix}
\usepackage{booktabs}
\usepackage{algorithm}
\usepackage{algpseudocode}
\usepackage{mathtools}

\usepackage{graphicx}
\usepackage{times}
\usepackage{soul}
\usepackage{url}
\usepackage{caption}
\usepackage[caption=false,font=footnotesize]{subfig}
\usepackage[utf8]{inputenc}
\usepackage{amsmath,amssymb}
\usepackage{array}
\usepackage{multirow}
\usepackage{stmaryrd}
\usepackage[misc]{ifsym}
\usepackage{tabularx}
\usepackage{xcolor}

\newcommand{\mhsa}{\operatorname{MHSA}}
\newcommand{\sa}{\operatorname{SA}}

\newcommand{\spec}{\text{spec}}
\newcommand{\spat}{\text{spat}}

\newcommand{\val}[2]{$#1\pm#2$}
\newcommand{\valb}[2]{$\mathbf{#1\pm#2}$}

\newcommand{\cmark}{\ding{51}}
\newcommand{\xmark}{\ding{55}}

\newacronym[plural=CPs]{cp}{CP}{Coalescent Projection}

\begin{document}
\let\WriteBookmarks\relax
\def\floatpagepagefraction{1}
\def\textpagefraction{.001}

\shorttitle{Cross-Domain Few-Shot Learning for Hyperspectral Image Classification Based on Mixup Foundation Model}  
\title[mode = title]{Cross-Domain Few-Shot Learning for Hyperspectral Image Classification Based on Mixup Foundation Model}
\shortauthors{N. Paeedeh et~al.}

\author[a]{Naeem Paeedeh}[orcid=0000-0001-9891-3294]
\ead{naeem.paeedeh@adelaide.edu.au}
\author[a,b]{Mahardhika Pratama}[orcid=0000-0001-6531-5087]
\ead{dhika.pratama@adelaide.edu.au}
\cormark[1]
\author[b]{Ary Shiddiqi}[orcid=0000-0002-8762-3141]
\ead{ary.shiddiqi@its.ac.id}
\author[a]{Zehong Cao}[orcid=0000-0003-3656-0328]
\ead{jimmy.cao@adelaide.edu.au}
\author[c]{Mukesh Prasad}[orcid=0000-0002-7745-9667]
\ead{Mukesh.Prasad@uts.edu.au}
\author[d]{Wisnu Jatmiko}[orcid=0000-0002-0530-7955]
\ead{wisnuj@cs.ui.ac.id}

\affiliation[a]{
            organization={School of Computer Science and IT, Adelaide University}, 
            city={Adelaide},
            postcode={5095}, 
            state={South Australia},
            country={Australia}
            }

\affiliation[b]{
            organization={Department of Informatics, Institut Teknologi Sepuluh Nopember}, 
            city={Surabaya},
            postcode={60111}, 
            state={East Java (Jawa Timur)},
            country={Indonesia}
            }

\affiliation[c]{
            organization={University of Technology, Sydney, Australia},
            city={Sydney},
            postcode={2007}, 
            state={New South Wales (NSW)},
            country={Australia}
            }

\affiliation[d]{
            organization={Faculty of Computer Science, University of Indonesia}, 
            city={Depok},
            postcode={16424}, 
            state={West Java (Jawa Barat)},
            country={Indonesia}
            }
            
\cortext[cor1]{Corresponding Author.}

\begin{abstract}
Although cross-domain few-shot learning (CDFSL) for hyper-spectral image (HSI) classification has attracted significant research interest, existing works often rely on an unrealistic data augmentation procedure in the form of external noise to enlarge the sample size, thus greatly simplifying the issue of data scarcity. They involve a large number of parameters for model updates, being prone to the overfitting problem. To the best of our knowledge, none has explored the strength of the foundation model, having strong generalization power to be quickly adapted to downstream tasks. This paper proposes the MIxup FOundation MOdel (MIFOMO) for CDFSL of HSI classifications. MIFOMO is built upon the concept of a remote sensing (RS) foundation model, pre-trained across a large scale of RS  problems, thus featuring generalizable features. The notion of coalescent projection (CP) is introduced to quickly adapt the foundation model to downstream tasks while freezing the backbone network. The concept of mixup domain adaptation (MDM) is proposed to address the extreme domain discrepancy problem. Last but not least, the label smoothing concept is implemented to cope with noisy pseudo-label problems. Our rigorous experiments demonstrate the advantage of MIFOMO, where it beats prior arts with up to $14\%$ margin. The source code of MIFOMO is open-sourced in \href{https://github.com/Naeem-Paeedeh/MIFOMO}{https://github.com/Naeem-Paeedeh/MIFOMO} for reproducibility and convenient further study.   
\end{abstract}



\maketitle
\section{Introduction}

A Hyperspectral image (HSI) plays an important role in the remote sensing (RS) field because it is, unlike RGB images, in the form of 3D data cubes and contains abundant spatial and spectral information revealing unique characteristics of various land covers \cite{Landgrebe2002HyperspectralID}. It is widely applied across different application domains such as mining, biology, and natural disasters, etc. 

The challenge of the HSI classification task is to assign a label to each pixel, which incurs high labeling costs. Early approaches to HSI classification problems make use of traditional machine learning methods \cite{Samaniego2008SupervisedCO,Melgani2004ClassificationOH} combined with manual feature extraction techniques. Nonetheless, such approaches are over-dependent on prior knowledge and suffer from significant bottlenecks in handling high-dimensional data such as HSIs. The advent of deep learning (DL) has revolutionized the RS field and has been successful across different RS problems, including the HSI classification problems. In contrast to traditional approaches, the DL methods feature an automatic feature engineering process, making it possible for deep and nonlinear features to be automatically extracted, learned, and classified \cite{Chen2014DeepLC}. Various models have been proposed to cope with the HSI classification problems, including CNN \cite{Lee2016GoingDW,Zhao2022SuperpixelGD,Zhang2018DiverseRC}, transformer \cite{Hong2021SpectralFormerRH,Wang2023DCNTDC}, graph neural network \cite{Liu2020CNNEnhancedGC,Dong2022WeightedFF}, etc. Nonetheless, the success of DL approaches is attributed to the availability of large datasets hindering their practical applications in the real-world HSI classification problems, imposing prohibitive labeling costs \cite{Paeedeh2024FewShotCI}. 

Such an issue has motivated the development of various methods to reduce the labeling cost problem. The semi-supervised learning (SSL) \cite{Yao2023PseudolabelBasedUS,Wei2023SemiSupervisedNA} and self-supervised learning (SSL) \cite{Li2024CrossDomainFH,Ma2023SelfSupervisedSG} methods have been proposed in the realm of the HSI classification problems. Although these approaches are capable of reducing the labeling cost to some extent, they call for a lot of samples. It is not well-suited to the HSI classification problem, where the cost of data collection is also expensive. This issue has led to the development of few-shot learning methods \cite{Snell2017PrototypicalNF,Sung2017LearningTC} for the HSI classification problems. The few-shot learning approach relies on an episodic learning procedure based on the so-called meta-learning approach. That is, the training process is divided into a number of episodes having very few samples, i.e., support and query sets. Such approaches allow to quickly generalize to unseen classes with limited samples because a model is trained to acquire meta-knowledge. Nevertheless, conventional few-shot learning methods aren't prepared to address the domain shift problem where the source and target domains are drawn from different distributions. This bottleneck is not in line with the fact that data collection of HSIs might be influenced by external factors, e.g., weather, lighting, sensor nonlinearities, etc., causing the issue of domain shifts. 

Cross-scene HSI classification methods \cite{Yang2016DomainAW,Ma2021CrossDatasetHI,Huang2022TwoBranchAA} are proposed to address the domain shift problem. This approach transfers knowledge of a label-rich domain known as the source domain to a label-scarce domain known as the target domain. The underlying assumption is that there are no labels at all in the target domain, and the presence of domain shifts between the source domain and the target domain. Usually, such approaches are driven by the domain adaptation (DA) method \cite{Huang2023CrosssceneWM} to align the distributional discrepancies between the source domain and the target domain. The key limitation of the cross-scene HSI classification lies in the closed-set assumption, where the source domain and the target domain share the same label space. Cross-Domain Few-Shot Learning (CDFSL) \cite{Zhang2022GraphIA,Paeedeh2024CrossDomainFL} aims to overcome this limitation, where the goal is to cope with the issue of data scarcity and domain shift simultaneously. It goes one step ahead of cross-scene HSI classification method,s where the label space between the source domain and the target domain can be completely different.    

The CDFSL \cite{Paeedeh2024CrossDomainFL} for HSI classification has grown rapidly, where numerous works have been proposed to deal with this problem. Existing works can be categorized into three groups: discrepancy-based methods \cite{Li2024SCFormerSC,Hu2023CrossDomainMU,Li2024SemanticGP,Dang2025CrossDomainFL,Qin2024FewShotLW,Zhang2022GraphIA,Qin2024CrossDomainFL,Feng2024CrossDomainFL}, adversarial-based methods \cite{Peng2023ConvolutionalTF,Shi2025MultibranchFT,Qin2024FewShotLW,Wang2022SpatialSpectralLD,Zhang2023CrossDomainSN,Ye2023AdaptiveDF,Cao2024SpatialSpectralSemanticCF,Wang2024DualBranchDA,Liu2025MultilevelPA,Shi2025MultiGaussianPM,Zhu2025FromIT} and contrastive-based methods \cite{Shi2025FewShotLB,Ye2024CrossDomainFL,Huang2023HyperspectralIC,Zhang2022CrossDomainFC,Li2024CrossDomainFH}. The discrepancy-based methods are designed to minimize distributional discrepancy between the source domain and the target domain using the discrepancy metrics. The adversarial-based methods utilize the adversarial domain adaptation technique \cite{Ganin2015DomainAdversarialTO}, playing the minmax game between the discriminator and generator. The contrastive-based methods apply the idea of contrastive learning to guarantee the compactness of intra-class data and the separability of inter-class data. Nonetheless, existing methods suffer from at least three drawbacks:
\begin{itemize}
    \item Existing methods apply unrealistic data augmentation procedures to enlarge the size of the target domain \cite{Wang2024DualBranchDA}. Such an approach greatly simplifies the issue of data scarcity but may be trapped in the out-of-distribution problem. The model performance may be compromised by the problem of sample diversity. In addition, the data augmentation procedure simply follows the RGB image augmentation strategy. The RGB image augmentation strategy might not be compatible with the HSI case. 
    \item Existing methods are prone to the issue of overfitting because they involve a large number of trainable parameters. The overfitting issue hinders the generalization aptitude of CDFSL models.  
    \item Very few works have explored the strength of hyperspectral foundation models in the realm of CDFSL for HSI classification problems. The foundation model possesses strong generalization potential, which can be quickly adapted to downstream tasks, thus aligning very well with the spirit of the CDFSL for HSI classification problems. To the best of our knowledge, only \cite{Chen2025SpectralDINODM} has pioneered the use of a foundation model in the context of CDFSL for HSI classification. However, this method still utilizes the foundation model of the RGB images, hindering their generalization for HSIs. Besides, we propose a new parameter-efficient fine-tuning (PEFT) method, namely coalescent projection (CP), whereas they \cite{Chen2025SpectralDINODM} are still based on the LoRA method.  
\end{itemize}

This paper proposes a mixup foundation model (MIFOMO) for CDFSL of HSI classification problems. MIFOMO is built upon a hyper-spectral foundation model, HyperSIGMA \cite{Wang2024HyperSIGMAHI}, pretrained in the HyperGlobal-450K dataset containing 450 K HSIs. This goes one step ahead of \cite{Chen2025SpectralDINODM}, still relying on the foundation model of RGB images, dinoV2. To remedy the issue of overfitting, we put forward a novel PEFT technique, namely CP \cite{Paeedeh2025CrossDomainFL}. That is, only CPs are trainable during the training process, leaving the backbone network frozen. This strategy also allows preservation of the strong generalization power of the foundation model. CP can be seen as a successor of plain prompt \cite{Wang2021LearningTP}, incurring significantly fewer parameters because it relies on a single learnable matrix. In addition, the CP technique is simpler to apply than the plain prompt because there is no need to select the prompt length. Another innovation of MIFOMO lies in the domain adaptation strategy, where the mixup domain adaptation approach \cite{Furqon2024MixupDA} is put into perspective. This strategy consists of two phases: source domain and intermediate domain, in which the episodic-based meta-learning approach is undertaken. Unlike direct knowledge transfer between the source and target domains, the intermediate domain functions as a bridge between the source and target domains, making seamless knowledge transfer possible because direct domain alignment may be too intricate to perform. Last but not least, the concept of label smoothing is incorporated to combat the noisy pseudo-label problem. That is, pseudo-labels are assigned to the query set of the target domain to prevent the label leakage problem. 

This paper proposes at least three contributions:
\begin{enumerate}
    \item This paper proposes the mixup foundation model (MIFOMO) for the CDFSL of HSI classification problems. This approach is unique in respect to state-of-the-art (SOTA) methods because it relies on the PEFT approach, i.e., existing works are based on the discrepancy-based methods, the adversarial-based methods, and the contrastive-based methods. It distinguishes itself from \cite{Chen2025SpectralDINODM} because of the hyperspectral foundation model and the CP technique. 
    \item The concept of CP is introduced to adapt the foundation model, namely HyperSIGMA, to the target domain. CP uses a single trainable matrix connecting the key and query projections in the attention map while leaving the backbone network frozen. CP offers very few trainable parameters and is easy to apply because it isn't dependent on any hyperparameters. We extend \cite{Paeedeh2025CrossDomainFL} designed for natural images to hyperspectral images here. 
    \item MIFOMO advances the mixup domain adaptation \cite{Furqon2024MixupDA} originally designed for unsupervised domain adaptation (UDA) scenarios, i.e., our concern is in the CDFSL of the HSI classification problem here. Moreover, the mixup domain adaptation method is generalized here using the idea of label smoothing to combat the noisy pseudo-label problem when meta-learning the target domain, while the episodic-based meta-learning approach is implemented in the source and intermediate domains. Besides, \cite{Furqon2024MixupDA} still works on the traditional architecture, whereas we implement such an idea in the context of a foundation model.  
\end{enumerate}

The remainder of this paper is structured as follows: Section 2 discusses related works, Section 3 outlines our method, MIFOMO, and Section 4 elaborates on experiments and analyses. Last but not least, some concluding remarks are drawn in the last section. 

\section{Related Works}
Cross-Domain Few-Shot Learning (CDFSL) for the Hyperspectral Image (HSI) classification problem is meant to address the problem of data scarcity and domain shift in the HSI classification problem. It aims to leverage discriminative information of the label-rich domain (source domain) to solve the HSI classification problem within the label-scarce domain (target domain). Compared to cross-scene HSI classification problems \cite{Yang2016DomainAW,Ma2021CrossDatasetHI,Huang2022TwoBranchAA},this problem features extreme domain discrepancies where the label spaces between the two domains might be completely disjoint. The CDFSL for HSI classification problems have grown rapidly where existing approaches can be divided into three categories based on how the domain alignment approach is performed: discrepancy-based methods \cite{Li2024SCFormerSC,Hu2023CrossDomainMU,Li2024SemanticGP,Dang2025CrossDomainFL,Qin2024FewShotLW,Zhang2022GraphIA,Qin2024CrossDomainFL,Feng2024CrossDomainFL}, adversarial-based methods \cite{Peng2023ConvolutionalTF,Shi2025MultibranchFT,Qin2024FewShotLW,Wang2022SpatialSpectralLD,Zhang2023CrossDomainSN,Ye2023AdaptiveDF,Cao2024SpatialSpectralSemanticCF,Wang2024DualBranchDA,Liu2025MultilevelPA,Shi2025MultiGaussianPM,Zhu2025FromIT} and contrastive-based methods \cite{Shi2025FewShotLB,Ye2024CrossDomainFL,Huang2023HyperspectralIC,Zhang2022CrossDomainFC,Li2024CrossDomainFH}. 
\subsection{Discrepancy-based method}
The discrepancy-based method measures the domain gap and minimizes it via a distance metric. \cite{Li2024SCFormerSC} utilizes the Wasserstein distance metric, \cite{Hu2023CrossDomainMU} applies the class covariance matrix, \cite{Li2024SemanticGP} uses the semantic information as complementary information of prototype construction, \cite{Dang2025CrossDomainFL} implements the CORAL strategy, \cite{Qin2024FewShotLW} makes use of the MMD metric. \cite{Zhang2022GraphIA} applies the graph mining strategy, \cite{Qin2024CrossDomainFL} designs the multiorder spectral interaction block, \cite{Feng2024CrossDomainFL} uses the knowledge distillation strategy. The aforementioned works directly minimize the discrepancy between two domains, which can be difficult in the case of extreme domain gaps and often result in negative transfer. Our approach here is based on the mixup domain adaptation, consisting of two stages: the source domain and the intermediate domain. The intermediate domain is crafted through the mixup mechanism and allows smooth knowledge transfer because it functions as a bridge between the source domain and the target domain. In addition, the direct domain adaptation method performs global domain alignments, ignoring the class alignment step, i.e., local. Note that the label spaces of the two domains don't overlap.  
\subsection{Adversarial-based Method}
The adversarial-based method follows an adversarial domain adaptation strategy \cite{Ganin2015DomainAdversarialTO} which introduces a domain discriminator detecting the origin of data samples, i.e., source or target domains. It plays a minmax game with the feature extractor, and the domain alignment stage is achieved when the feature extractor successfully fools the domain discriminator. \cite{Peng2023ConvolutionalTF} applies the vanilla adversarial domain adaptation method, \cite{Shi2025MultibranchFT} implements the conditional adversarial domain adaptation method, \cite{Qin2024FewShotLW} constructs an intermediate domain, \cite{Wang2022SpatialSpectralLD} devises the local and global adversarial domain adaptation method, \cite{Zhang2023CrossDomainSN} utilizes the collaborative adversarial network concept. \cite{Ye2023AdaptiveDF} inserts a varying weight to the conditional adversarial loss, \cite{Cao2024SpatialSpectralSemanticCF} uses a semantic-aware strategy for the domain discriminator, \cite{Wang2024DualBranchDA} implements the random sampling strategy to address the curse of dimensionality, and \cite{Liu2025MultilevelPA} applies the prototype-based conditional adversarial domain adaptation method. \cite{Shi2025MultiGaussianPM} proposes a cyclic domain triplet loss combining the adversarial domain adaptation technique and the contrastive learning strategy, while \cite{Zhu2025FromIT} offers a cyclic resemblance domain adversarial network. As with the discrepancy-based method, the adversarial-based method performs the global domain alignment strategy, excluding the class alignment step.

\subsection{Contrastive-based method}
The contrastive-based method relies on the contrastive learning strategy to attain alignments between the two domains. \cite{Shi2025FewShotLB} proposes the inter-domain contrastive loss function, \cite{Ye2024CrossDomainFL} puts forward the graph convolutional contrast module, \cite{Huang2023HyperspectralIC} puts into perspective the kernel triplet loss function, \cite{Zhang2022CrossDomainFC,Li2024CrossDomainFH} make use of the supervised contrastive loss function. Although the contrastive learning strategy is deemed feasible to achieve the class alignment stage, it requires the sample diversity, which is not served by the na\"ive data augmentation strategy, such as the Gaussian noise. Besides, we offer a new perspective in this paper where the PEFT strategy is proposed for the hyperspectral foundation model. 

\section{Preliminaries}
\subsection{Problem Definition}
The cross-domain few-shot learning (CDFSL) for the HSI classification problem is formulated as a learning problem of two distinct domains, namely the source domain $\mathcal{D}_{S}$ and the target domain $\mathcal{D}_{T}$, $\mathcal{D}_{S}\neq\mathcal{D}_{T}$. The source domain $\mathcal{D}_{S}$ is fully labeled with a label space $\mathcal{Y}_{S}$ while the target domain $\mathcal{D}_{T}$ consists of extremely few labeled samples with a label space $\mathcal{Y}_{T}$. The label space of the two domains are disjoint $\mathcal{Y}_{S}\cap\mathcal{Y}_{T}=\emptyset$ but the number of classes of the source domain is larger than the target domain $\mathcal{C}_{S}\geq\mathcal{C}_{T}$ to ensure sample diversity.

Given the source domain as an instantiation, the training process occurs in an episodic-based meta-learning manner. That is, each episode $\mathcal{T}_{S}$ is constructed from the support set $\mathcal{S}_{s}$ and the query set $\mathcal{Q}_{s}$ $\mathcal{T}_{s}=\{\mathcal{S}_{s},\mathcal{Q}_{s}\}$. The support set $\mathcal{S}_{s}$ is formed in the $N$-way $K$-shot setting $\mathcal{S}_{s}=\{(x_{i},y_{i})\}_{i=1}^{N\times K}$ where $N$ refers to the number of classes randomly selected from the training data while $K$ constitutes the number of samples per classes. On the other hand, the query set $\mathcal{Q}_{s}$ is formulated in the $N$-way $Q$-shot setting $\mathcal{Q}_{s}=\{(x_{i},y_{i})\}_{i=1}^{N\times Q}$. The same case happens in the target domain except the query set $\mathcal{Q}_{t}$ is unlabeled $\mathcal{Q}_{t}=\{(x_{i})\}_{i=1}^{N_{q}}$ and there exist extremely few labeled samples as the support set $\mathcal{S}_{t}=\{(x_i,y_i)\}_{i=1}^{N\times K}$. That is, the pseudo-labeling strategy is implemented for $\mathcal{Q}_{t}$ and no data augmentation is executed to perform the episodic-based meta learning in the target domain.

Our network $f_{\theta}=h_{\phi}\circ g_{\psi}$ is composed of a feature extractor extracting the latent vectors $g_{\psi}:\mathcal{X}\rightarrow\mathcal{Z}$ and a classifier converting the latent vectors to the category labels $h_{\phi}:\mathcal{Z}\rightarrow\mathcal{Y}$. $\theta=\{\phi,\psi\}$ is the network parameter. The parameter-efficient fine-tuning (PEFT) strategy is adopted, where the backbone network $f_{\theta}$ is kept frozen during the training process and utilizes small external trainable parameters for adaptations.  

\subsection{Mixup Strategy}
Our approach, MIFOMO, leverages the mixup strategy to enhance the sample's diversity, deemed as the consistency-based regularization strategy, forcing the network to output the same predictions for interpolated samples \cite{Mai2019MetaMixUpLA}. It has been shown to meet a lower bound of the Lipschitz constant of the gradient of the neural network. In the realm of the HSI classification problem using the deep learning approach, this strategy can be committed in both the image level or the embedding level and generates an augmented sample pair based on a linear interpolation strategy. Given two distinct HSIs $(x_i,x_j)$ and their corresponding labels $(y_i,y_j)$, the mixup method is described as follows:
\begin{equation}\label{mixup}
    \begin{split}
        \tilde{x}&=\lambda x_{i}+(1-\lambda)x_j\\
        \tilde{y}&=\lambda y_{i}+(1-\lambda)y_{j},
    \end{split}
\end{equation}
where $\lambda$ is a mixup ratio playing a vital role in controlling the influence of the two samples. It is randomly drawn from the Beta distribution $\beta(\alpha,\alpha)$. $\alpha\in(0,\infty)$ is the parameter of the Beta distribution known as the confidence level. The mixup samples can be learned akin to the original samples, i.e., via the cross-entropy loss, and thereby improving the trained network's robustness. 
\subsection{Prototypical Network}
MIFOMO is built upon the idea of prototypical network \cite{Snell2017PrototypicalNF} using the prototypes to classify the query samples. The prototypes are the centers of the samples of each class, quantified as follows:
\begin{equation}\label{eq:prototypes}
    \mu_{c}=\frac{1}{N_{c}}\sum_{i=1}^{N_{c}}g_{\psi}(x_{i}),
\end{equation}
where $N_{c}$ denotes the number of samples of the $c-th$ class. The classification process is carried out in a distance-based manner as follows:
\begin{equation}
    P(y_{j}=c|x_{j}\in\mathcal{Q})=\frac{\exp{\bigl(-D(\mu_{c},z_{j})\bigr)}}{\sum_{c=1}^{C}\exp{\bigl(-D(\mu_{c},z_{j})\bigr)}}.
\end{equation}
In the realm of episodic-based meta-learning, the prototypes are concluded from the support set, while the few-shot learning loss is calculated using the query set in each episode. The few-shot learning loss is mathematically written as follows:
\begin{equation}\label{FSL_loss}
\mathcal{L}_{\text{fsl}}=\mathbb{E}_{\mathcal{S},\mathcal{Q}}\Bigl[-\sum_{(x,y)\in\mathcal{Q}}\log{\bigl(P(y_{j}=c|x_{j})\bigr)}\Bigr].
\end{equation}
This process repeats for all episodes. In addition, MIFOMO applies this strategy for the source and intermediate domains. 

\begin{algorithm}[!t]
    \caption{Pseudo-code of the MIFOMO}
    \label{alg:MIFOMO}
    \begin{algorithmic}[1]
        \State $\triangleright$ Source domain
        \State \textbf{Input:} Source dataset $\mathcal{D}_{S}$
        \State \textbf{Output:} Trained model
        
        \For{episode id $\in \{1,\dots,E_{\text{s}}\}$}
            \State Obtain the support set $\mathcal{S}_{s}$ and query set $\mathcal{Q}_{s}$
            \State Calculate the prototypes from the $\mathcal{S}_{s}$  \Comment{Eq.~\ref{eq:prototypes}}
            \State Choose pairs of samples from the $\mathcal{Q}_{s}$
            \State Mix the embeddings $\tilde{g}_{\psi}(x)$ and labels $\tilde{y}$ \Comment{Eq.~\ref{eq:mixup,source}}
            \State Calculate the $\mathcal{L}_{\text{fsl}}^{S}$ and $\mathcal{L}_{\text{mx}}$ \Comment{Eq.~\ref{eq:loss_mixup_source}}
            \State Calculate and optimize the $\mathcal{L}_{S}=\mathcal{L}_{\text{fsl}}^{S}+\mathcal{L}_{\text{mx}}^{S}$ \Comment{Eq.~\ref{eq:loss_total_source}}
        \EndFor
        
        \State $\triangleright$ Intermediate domain
        \State \textbf{Input:} Trained model from the source domain, Source dataset $\mathcal{D}_{S}$, and target dataset $\mathcal{D}_{T}$
        \State \textbf{Output:} Trained model
        
        \For{episode id $\in \{1,\dots,E_{\text{intr}}\}$}
            \State Obtain the support sets $\mathcal{S}_{s}$ and $\mathcal{S}_{t}$, and query sets $\mathcal{Q}_{s}$ and $\mathcal{Q}_{t}$
            \For{each $\mathcal{\overline{S}}_{S^T}, \mathcal{\overline{Q}}_{S^T}$ by splitting the $\mathcal{S}_{t}$}
                \State Train the model like the source domain \Comment{Eq. \ref{eq:mixup_for_pseudo-labels}, \ref{eq:loss_mixup_for_pseudo-labels},and \ref{eq:loss_total_mixup_for_pseudo-labels}}
            \EndFor
            \State Assign the pseudo-labels to the $\mathcal{Q}_{t}$
            \For{each mini-batch $B$ in $\mathcal{Q}_{t}$}
                \State Concatenate $\mathcal{S}_{t}$ and $B$
                \State Apply label-smoothing \Comment{Eq. \ref{eq:label-smoothing}}
                \State Discard the support set and keep the query set mini-batch
            \EndFor
            \State Find the top-k query samples with the highest confidence scores and add them to the support set as $\hat{\mathcal{D}}_{T}$
            \For{each episode id $\in \{1,\dots,E_{\text{intr}}\}$}
                \State Obtain the support sets $\mathcal{S}_{s}$ and $\hat{\mathcal{S}}_{t}$, and query sets $\mathcal{Q}_{s}$ and $\hat{\mathcal{Q}}_{t}$
                \State Calculate the prototypes from the $\mathcal{S}_{s}$ and $\hat{\mathcal{S}}_{t}$
                \State Choose pairs of samples from the $\mathcal{Q}_{s}$ and $\hat{\mathcal{Q}}_{t}$
                \State Mix the pairs of source and target inputs, embeddings, and labels to obtain the $\tilde{x}, \tilde{g}_{\psi}(x)$, and $\tilde{y}$ \Comment{Eq.~\ref{eq:mix_intermediate_inputs_and_labels}, \ref{eq:mix_intermediate_embeddings_and_labels}}
                \State Calculate and optimize the $\mathcal{L}_{\text{inter}}$ \Comment{Eq.~\ref{eq:loss_intermediate}}
            \EndFor
            \State Update the $\lambda_2^n$ \Comment{Eq.~\ref{eq:q_for_lambda}, \ref{eq:lambda_update}}
        \EndFor
        
    \end{algorithmic}
\end{algorithm}

\section{Methodology}
MIFOMO is constructed under a hyper-spectral foundation model, namely HyperSIGMA \cite{Wang2024HyperSIGMAHI}, pretrained using the Hyper-Global-450K dataset by means of the masked image modeling technique. HyperSIGMA is a vision transformer (ViT)-based foundation model \cite{Dosovitskiy2020AnII}, and its innovation lies in the proposal of the sparse sampling attention (SSA) method instead of the self-attention mechanism. Unlike \cite{Chen2025SpectralDINODM}, using the natural-images foundation model, HyperSIGMA is meant specifically to handle HSIs. We propose a coalescent projection (CP) method as a parameter-efficient fine-tuning (PEFT) technique while leaving the HyperSIGMA backbone completely frozen. The CP method establishes a single learnable matrix that embraces the query-key interactions. We put forward the mixup strategy where the episodic-based meta-learning phase occurs in the source and intermediate domains. The intermediate domain is crafted from the mixup samples between the source and target domains and functions as a bridge between the two domains for the sake of seamless knowledge transfer. 

\subsection{Network Structure of HyperSIGMA}
\begin{figure*}[t!]
    \centering
    \includegraphics[width=0.8\linewidth]{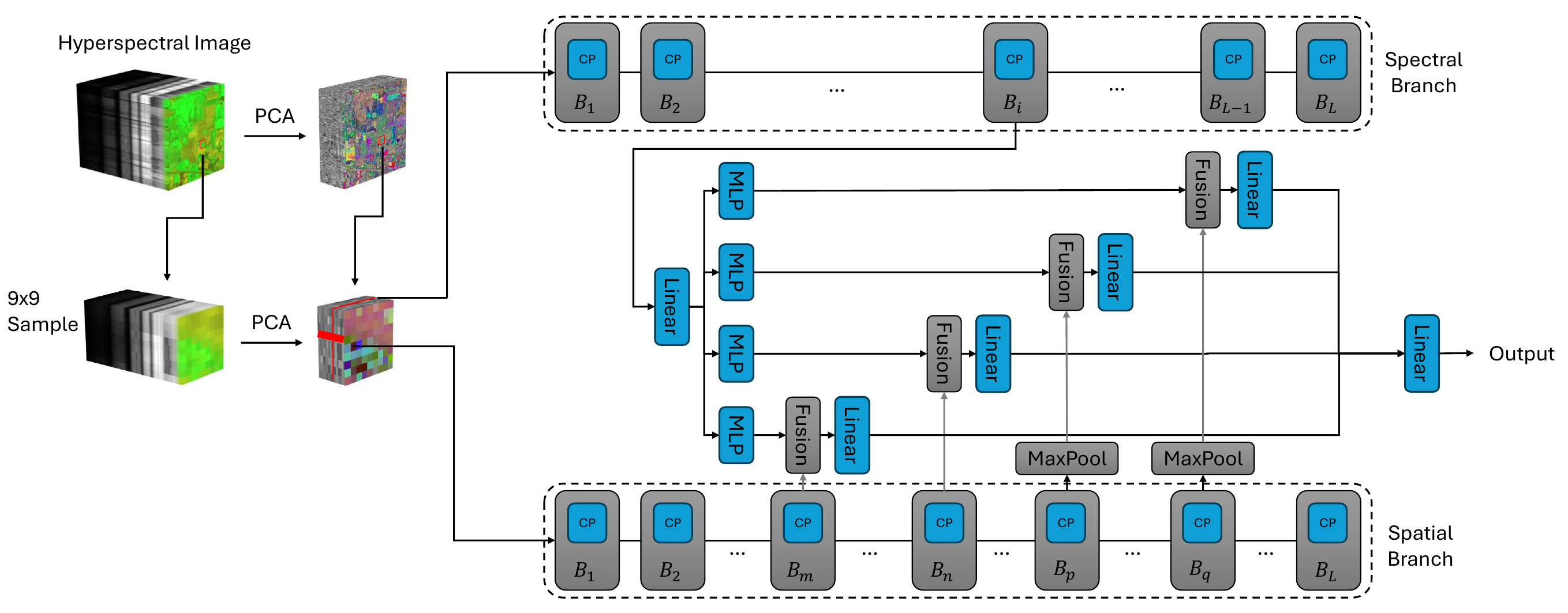}
    \caption{The architecture of our model. The blue blocks contain learnable parameters.}
    \label{fig:Architecture}
\end{figure*}

HyperSIGMA \cite{Wang2024HyperSIGMAHI} comprises two sub-networks: SpatialNetwork and SpectralNetwork, where spatial and spectral features are ultimately fused to enhance the representation of the extracted features. Fig. \ref{fig:Architecture} visualizes the architecture of MIFOMO. Both networks are based on the ViT backbone. However, the spatial tokenization is extended to the spectral domain, generating spectral tokens by embedding channels. That is, a 3-D HSI cube $x_0\in\Re^{H\times W\times C}$ is aggregated through the average clustering technique along the channel dimension, leading to $X^{'}\in\Re^{H\times W\times N_{\spec}}$ where $N_{\spec}$ is the desired token number. The dimensional permutation and flattening are applied to reshape $X^{'}$ into a 2-D matrix of $\Re^{N_{\spec}\times(H.W)}$. It is then embedded through a linear mapping and generates $X\in^{N_{\spec}\times D}$, being akin to the spatial patch embedding in normal ViT.  

The ViT building block consists of the self-attention layer and the feed-forward component, and splits an image into non-overlapping patches mapped to a $1-D$ vector known as the patch embedding mechanism. The tokens are added with learnable positional embeddings $P$ and then fed to a series of transformer blocks $f=\{f_{1},...,f_{d}\}$ where $d$ is the number of transformer blocks. 
\begin{equation}
    \begin{split}
        U_{0}&=X+P, X\in\Re^{N\times D}, E\in\Re^{N\times D}\\
        U_{i}&=f_{i}(U_{i-1}), U_{i}\in\Re^{N\times D}\\
        Z&=\operatorname{LN}(U_{d}), Z\in\Re^{N\times D},
    \end{split}
\end{equation}
where $\operatorname{LN}(.)$ stands for the layer normalization. $U_{i}$ is processed by a multi-head self-attention layer and a feed-forward block as follows:
\begin{equation}
\begin{split}
    U_{i-1}^{'}&=\mhsa\bigl(\operatorname{LN}(U_{i-1})\bigr)+U_{i-1}\\
    U_{i} &= \operatorname{FFN}\bigl(\operatorname{LN}(U^{'}_{i-1})\bigr)+U_{i-1}^{'},
\end{split}
\end{equation}
where $FFN(.)$ and $\mhsa(.)$ respectively denote the feed-forward block containing two linear layers and the multi-head self-attention block. $\mhsa(.)$ comprises several self-attention layers $\sa(.)$ and operates in parallel. 
\begin{equation}\label{SA}
    \sa(U)=\operatorname{Softmax}(\frac{QK^{T}}{\sqrt{D^{'}}})V,
\end{equation}
where $Q\in\Re^{N\times D^{'}}=UW_{Q}$, $K\in\Re^{N\times D^{'}}=UW_{K}$, $V\in\Re^{N\times D^{'}}=UW_{V}$ are the query, key and value. The output of each $\sa(.)$ module is concatenated and linearly projected to generate the output of $\mhsa(.)$ module. 
\begin{equation}
    \mhsa(U)=\bigl[\operatorname{Concat}\bigl(\sa_{1}(U),\dots,\sa_{h}(U)\bigr)W\bigr]^{T},
\end{equation}
 where $h$ is the number of heads and $W\in\Re^{hD^{'}\times D}$ is a linear projector to recover the original embedding dimension $D$. We choose to retain the SA mechanism for all layers rather than using the SSA method because the SSA method is not applied to all layers in HyperSIGMA. It includes replacing the original SA with SSA, complicating the fine-tuning process.

 HyperSIGMA possesses a dual-branch network structure that produces spectral and spatial features, respectively. The two features are fused to enhance their representations. To this end, the spectral enhancement module (SEM) is designed to enhance spatial features with spectral information. Suppose that we have a spatial feature $Z_{\spat}\in\Re^{H^{'}\times W^{'} \times D}$ and a spectral feature $Z_{\spec}\in\Re^{N_{\spec}\times D}$, a linear layer is applied to reduce their dimensions to $D_{1}$. We further reduce the spatial dimension of $Z_{\spec}$ via average aggregation to produce a $1D$ vector $Z^{'}\in\Re^{N_{\spec}}$. Lastly, another linear mapping is implemented to align the dimension of $Z^{'}$ with the channel number of $Z_{\spat}$. The spectral enhancement process is formalized:
 \begin{equation}
     Z^{*}=(1+Z^{'})*Z_{\spat},
 \end{equation}
 where $*$ is the channel-wise product. 

\subsection{Coalescent Projection}
\begin{figure*}[t!]
    \centering
    \includegraphics[width=0.8\linewidth]{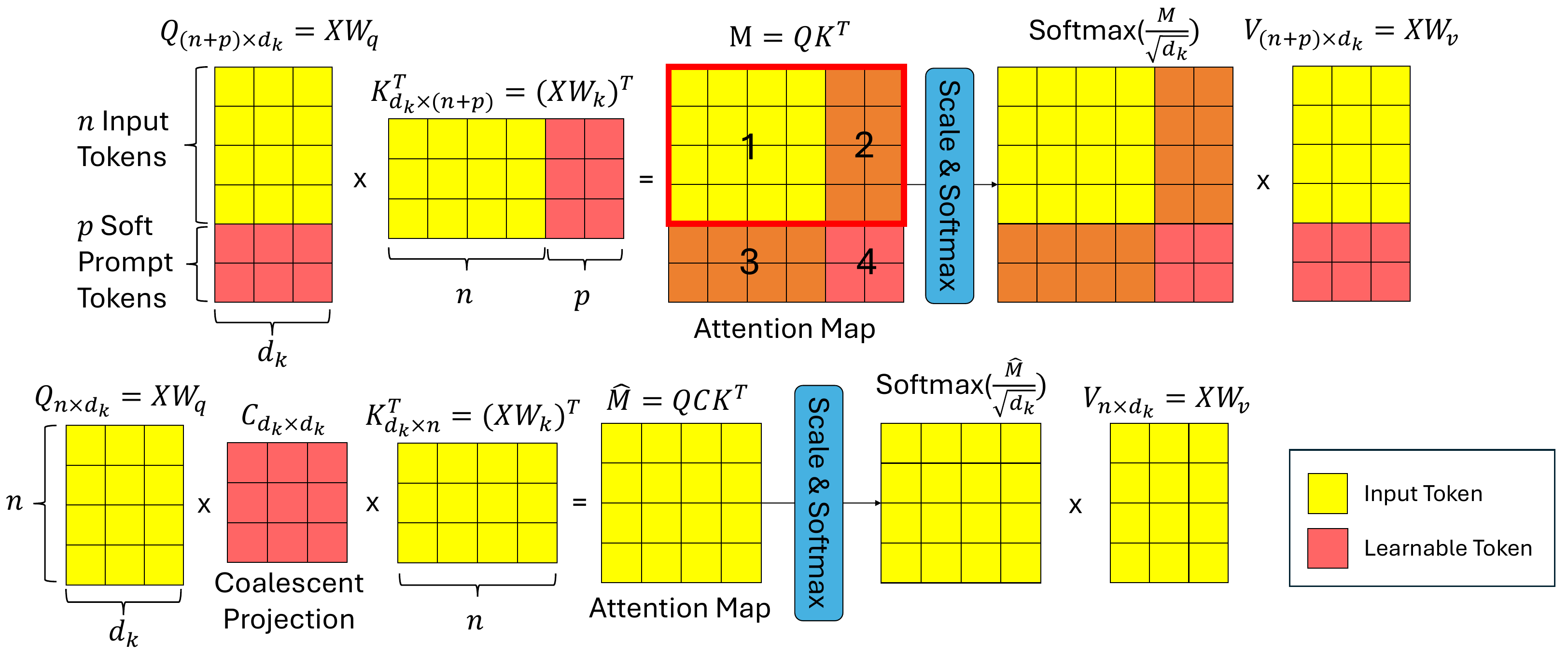}
    \caption{The calculation of the soft prompts on top and \Gls{cp} at the bottom, in the attention module.}
    \label{fig:PEFT_methods}
\end{figure*}

 MIFOMO is equipped with a coalescent projection (CP) seen as a generalization of the soft prompt \cite{Wang2021LearningTP}. We start our discussion by introducing the attention map, being a crucial part of the self-attention mechanism as per \eqref{SA}. It can be decomposed into 4 blocks.
 \begin{equation}
     A=\frac{QK^{T}}{\sqrt{D^{'}}}=\begin{bmatrix}
        Q_{1} K_{1}^{T} & Q_{1} K_{p}^{T} \\
        Q_{p}  K_{1}^{T} & Q_{p} K_{p}^{T} \\
    \end{bmatrix} = \\
    \begin{bmatrix}
        A_{1,1} & A_{1,p} \\
        A_{p,1} & A_{p,p} \\
    \end{bmatrix}.
 \end{equation}
 From \cite{Basu2023StrongBF}, we understand that the most crucial part of the attention map calculation is $A_{1,1}$ because the prompt can only affect $A_{1,1}$ and $A_{1,p}$ due to the presence of the softmax function. Since the prompt is injected in every layer to correct the influence of the prompt in the previous layer, the bottom part of the attention map $A$ is deemed redundant. Figure \ref{fig:PEFT_methods} exhibits the self-attention mechanism of the plain prompt \cite{Wang2021LearningTP} and the CP in this paper.  

The most straightforward approach to tune $A_{1,1}$ is to learn the key and query projection $W_{Q}, W_{K}$, but this imposes too many parameters, leading to the overfitting problem. In addition, the backbone network should be frozen to enjoy the generalization power of the foundation model. We propose to insert a single learnable matrix between the query and key to steer their directions. That is, a single learnable matrix $C$ is appended between the key and query matrices.
\begin{equation}
    \sa(U)=\operatorname{Softmax}\bigl(\frac{QCK^{T}}{\sqrt{D^{'}}}\bigr)V,
\end{equation}
 where $C$ is the only learnable component. It unifies projected key and query matrices to combine them into a single concept, namely attention map. This is enabled by the coalescent projection rather than via two projections. Suppose that $\Delta W_{q}$ and $\Delta W_{k}$ are the ideal changes of the query and key, the CP can be initialized close to the identity matrix as follows:
 \begin{equation}
     C=I + \Delta W_k^{T} {W_k^{T}}^\dagger + W_q^\dagger \Delta W_q + W_q^\dagger \Delta W_q \Delta W_k^{T} {W_k^{T}}^\dagger.
 \end{equation}
 The CP concept is applied separately in each head, making it possible for each head to be handled without interference. In addition, the CP method is applied in each layer. 

 Compared to the prompt method \cite{Wang2021LearningTP} inserting external trainable parameters to the patch embedding, the CP concent has at least three clear advantages: 1) the CP method doesn't have any specific hyper-parameter whereas in the prompt method one has to select an appropriate length of the prompt token; 2) the prompts are shared across all heads hampering model's generalization due to the gradient's interference whereas the CP concept handles each head independently; 3) the prompt increases the size of patch dimension consuming the context window and memory whereas the CP technique doesn't manipulate the calculation size from that of the original attention calculation; 4) the CP method can change the order of the original input tokens whereas the prompt doesn't have such aptitude. On the other hand, the CP method also imposes significantly fewer parameters than the attention scale \cite{Basu2023StrongBF}. This is especially important for high-resolution images such as HSIs. In addition, the attention scale method is not applicable to different input tokens.

\subsection{Mixup Domain Adaptation Strategy}
Motivated by \cite{Furqon2024MixupDA}, the mixup domain adaptation strategy is proposed to address the domain shift problem in the CDFSL for the HSI classification problem. This strategy consists of two phases, namely the source domain and the intermediate domain. Note that the episodic-based meta-learning strategy is implemented rather than the conventional training paradigm.  

\subsubsection{Source Domain}
\begin{figure*}[t!]
    \centering
    \includegraphics[width=0.8\linewidth]{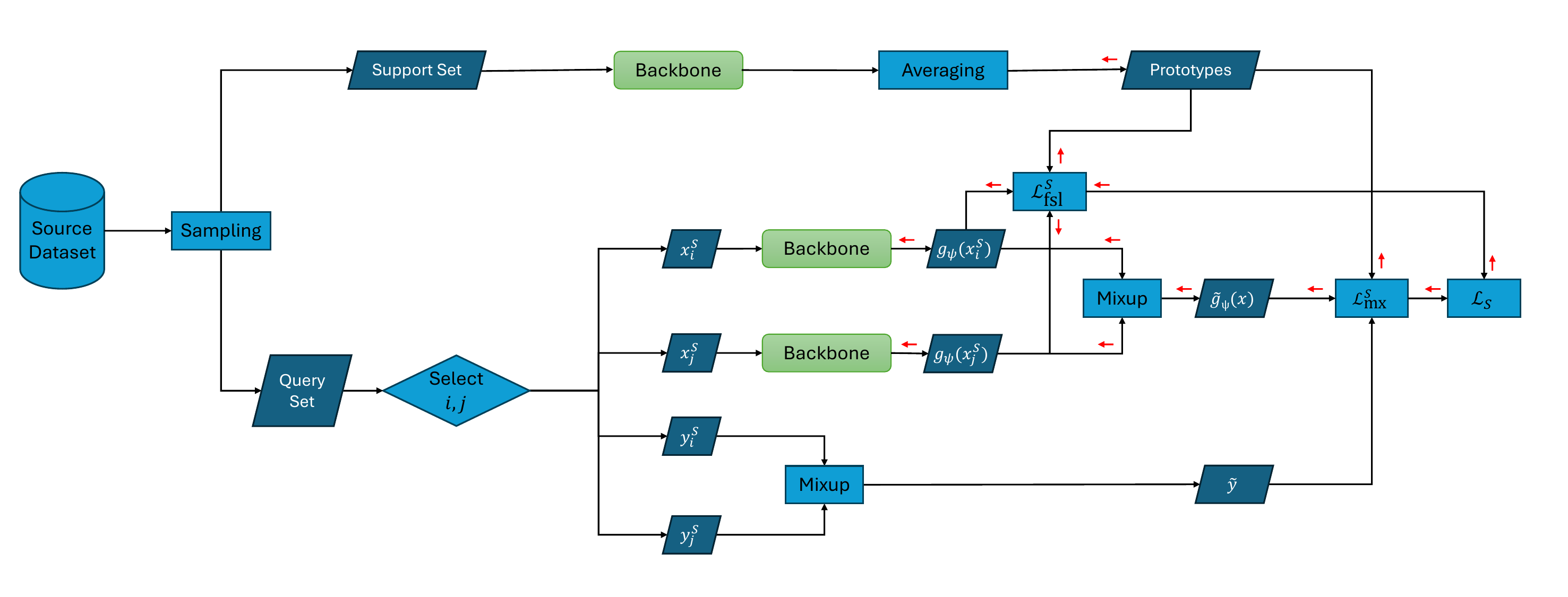}
    \caption{The flowchart of the single iteration of source domain training. The red arrows indicate the backpropagated gradients.}
    \label{fig:Flowchart_Source}
\end{figure*}

The mixup strategy is applied in the source domain to improve the samples' diversity. Because the source domain already contains plentiful samples, the mixup strategy is only applied in the feature space. That is, it is similar to \eqref{mixup}, yet the mixup mechanism is carried out in the embedding space. In addition, the embedding space is usually lower-dimensional than the image space. 
\begin{equation}
    \label{eq:mixup,source}
    \begin{split}
        \tilde{g}_{\psi}(x)&=\lambda_{1} g_{\psi}(x_{i}^{S})+(1-\lambda_{1})g_{\psi}(x_{j}^{S})\\
        \tilde{y}&=\lambda_{1}y_{i}^{S}+(1-\lambda_{1})y_{j}^{S},
    \end{split}
\end{equation}
where $\lambda_{1}$ is the mixup ratio drawn from the Beta distribution. Note that the mixup mechanism produces a convex combination of a pair of samples $(x_{i},y_{i})$ and $(x_{j},y_{j})$ from the query set. The augmented samples can be learned in a supervised manner via the cross-entropy loss function as per \eqref{FSL_loss}.
\begin{equation}
\label{eq:loss_mixup_source}
\mathcal{L}_{\text{mx}}^{S}=\mathbb{E}_{(x,y)\backsim\mathcal{D}_{S}}\Bigl[l\Bigl(f_{\phi}\bigl(\tilde{g}_{\psi}(x)\bigr),\tilde{y}\Bigr)\Bigr],
\end{equation}
where $l(.)$ is the cross-entropy loss function. The mixup loss function also functions as a consistency regularization, where the network is forced to produce consistent outputs to the interpolated samples. The total loss function of the source domain is defined as follows:
\begin{equation}
\label{eq:loss_total_source}
\mathcal{L}_{S}=\mathcal{L}_{\text{fsl}}^{S}+\mathcal{L}_{\text{mx}}^{S}
\end{equation}
 (15) is solved using the episodic-based meta-learning strategy for both original and mixup samples. That is, the loss is calculated from the query set of both original and mixup samples. This strategy enhances the robustness of the network because it aids the network in reserving the uncharted latent space uncovered by the original samples \cite{Paeedeh2025CrossDomainFL}. That is, it increases the complexity of the decision boundaries. Fig.~\ref{fig:Flowchart_Source} shows the flowchart of the source domain training phase.
 
\subsubsection{Intermediate Domain}
The mixup domain strategy involves creating an intermediate domain via the mixup mechanism between source-domain and target-domain samples. The intermediate sample is designed to enable seamless knowledge transfer between the source and target domains and to bridge the two domains. Unlike directly learning the discrepancy between the domains, having a problem in the case of large domain gaps, the intermediate domain allows the network to learn the representation of the source domain to the target domain smoothly. By extension, the mixup mechanism occurs in both image and embedding levels to enforce strong consistency regularization since both image and embedding levels should generate similar outputs. 

The first step is performed by assigning high-quality pseudo-labels to the target domain's query set. Next, the network is trained on the intermediate domain.
To assign pseudo-labels, the network is trained only on the support set of the target domain. This is done by sampling smaller support set $\mathcal{\overline{S}}_{S^T}=\{(x_{i},y_{i})\}_{i=1}^{N \times K_s}$ and query sets $\mathcal{\overline{Q}}_{S_T}=\{(x_{i},y_{i})\}_{i=1}^{N \times K_q}$ per episode as subsets of the current support set and using mixup in the feature space, as in the source-domain phase, where $K_s$ and $K_q$ are the number of support and query subset shots, respectively, and $K_s + K_q = K$.

The mixup is done in the embedding space as follows:

\begin{equation}
    \label{eq:mixup_for_pseudo-labels}
    \begin{split}
        \tilde{g}_{\psi}(x)&=\lambda_{1} g_{\psi}(x_{i}^{T})+(1-\lambda_{1})g_{\psi}(x_{j}^{T})\\
        \tilde{y}&=\lambda_{1}y_{i}^{S}+(1-\lambda_{1})y_{j}^{S},
    \end{split}
\end{equation}

and the mixup loss function is expressed as follows:

\begin{equation}
    \label{eq:loss_mixup_for_pseudo-labels}
    \mathcal{L}_{\text{mx}}^{\hat{T}}=\mathbb{E}_{(x,y)\backsim\mathcal{D}_{\hat{T}}}\Bigl[l\Bigl(f_{\phi}\bigl(\tilde{g}_{\psi}(x)\bigr),\tilde{y}\Bigr)\Bigr],
\end{equation}

\begin{equation}
    \label{eq:loss_total_mixup_for_pseudo-labels}
    \mathcal{L}_{\hat{T}}=\mathcal{L}_{\text{fsl}}^{\hat{T}}+\mathcal{L}_{\text{mx}}^{\hat{T}}.
\end{equation}

\label{pseudo-labeling}
The pseudo-label is generated based on a small number of samples of the support set of the target domain. That is, the prototypes are calculated based on the support set of the target domain $\mathcal{S}_{t}=\{(x_i,y_i)\}_{i=1}^{N\times K}$. However, these pseudo-labels are noisy because of the domain shift problem, i.e., the network is not yet trained to the target domain, and the target domain shares no overlapping label space with the source domain. To this end, the label smoothing strategy \cite{Zhou2003LearningWL} is utilized and enforces consistent predictions of adjacent samples. That is, neighboring samples must be labeled similarly or be smooth. A graph $G=(V,E)$ is constructed where $E$ is the nodes of the graph defined by all images in $\mathcal{Q}_{t}$ and $E$ is the edges weighted by an adjacency matrix $A\in\Re^{N_{Q}\times N_{Q}}$ whose elements are derived from the distance between two images in the feature space $a_{i,j}=\exp{(-\frac{(g_{\psi}(x_{i})-g_{\psi}(x_j))^{2}}{2\sigma^{2}})},j\neq i, a_{i,i}=0$. $\sigma$ is the spread of the latent space. The adjacency matrix is normalized as per $\hat{A}=D^{-1/2}AD^{-1/2}$ in which $D$ is a diagonal matrix whose diagonal element stands for the sum of the $i-th$ row of $A$. The label propagation is iterated until convergence $\mathcal{F}^{*}$. 
\begin{equation}
    \mathcal{F}_{t+1}=\alpha\hat{A}\mathcal{F}_{t}+(1-\alpha)\hat{Y},
\end{equation}
where $\mathcal{F}\in\Re^{N_{Q}\times\mathcal{C}_{T}}$ denotes the label assignment matrix and $\hat{Y}\in\Re^{N_{Q}\times\mathcal{C}_{T}}$ stands for the initial label assignment matrix assigned as the one-hot encoded form of the pseudo-label. $\alpha\in[0,1]$ is a hyper-parameter controlling the trade-off between the smoothness constraint and the fitting constraint. This has a closed-loop solution as follows:
\begin{equation}
    \label{eq:label-smoothing}
    \mathcal{F}^{*}=(I-\alpha\hat{A})^{\dagger}\hat{Y},
\end{equation}
where $\dagger$ is the pseudo-inverse symbol. The final pseudo-label is obtained from the label assignment matrix $\hat{y}_{i}=\arg\max_{c\in\mathcal{C}_{T}}\mathcal{F}_{i}^{*}$. Note that this mechanism does not leak any true class label. 

Next, the pseudo-labeling step is committed to unlabeled samples of the query set of the target domain thus generating pairs of samples $\mathcal{Q}_{t}=\{(x_{i},\hat{y}_{i})\}_{i=1}^{N\times Q}$, where the number of pseudo-labeled samples per class is set similarly to that of the source domain and $\hat{y}_{i}$ is a pseudo-label of the target domain samples $x_{i}$. Once concluding the pseudo-labeling step, the intermediate domain $\mathcal{\tilde{D}}$ is established in the image and feature spaces as follows:
 \begin{equation}
    \label{eq:mix_intermediate_inputs_and_labels}
     \begin{split}
         \tilde{x}&=\tilde{\lambda}_{2}x_{i}^{S}+(1-\tilde{\lambda}_{2})x_{j}^{T}\\
         \tilde{y}&=\tilde{\lambda}_{2}y_{i}^{S}+(1-\tilde{\lambda}_{2})\hat{y}_{j}^{T}
     \end{split}
 \end{equation}
\begin{equation}
    \label{eq:mix_intermediate_embeddings_and_labels}
    \begin{split}
        \tilde{g}_{\psi}(x)&=\tilde{\lambda}_{2} g_{\psi}(x_{i}^{S})+(1-\tilde{\lambda}_{2})g_{\psi}(x_{j}^{T})\\
        \tilde{y}&=\tilde{\lambda}_{2}y_{i}^{S}+(1-\tilde{\lambda}_{2})\hat{y}_{j}^{T}.
    \end{split}
\end{equation}
The mixup ratio $\lambda_{2}$ plays a vital role in steering the relationship of the mixup samples to the source and target domains, determining knowledge transfer. We follow the progressive mixup strategy \cite{Zhu2023ProgressiveMF}, where the Wasserstein distance $d(.)$ is applied to measure the discrepancies between the two domains, i.e., intermediate to source and intermediate to target. Initially, the intermediate domain is set to be close to the target domain.
The mixup ratio is gradually adjusted to bring the intermediate domain closer to the source domain, thereby achieving domain alignment. This setting reduces large domain gaps because the network is trained to master the target domain and subsequently the source domain, thus ensuring seamless knowledge transfer. This is attained by introducing the weighting factor $q$ to portray the similarity between the two domains. 
\begin{equation}
    \label{eq:q_for_lambda}
    q=\exp{\Bigl(\frac{-d(\mathcal{\tilde{D}},\mathcal{D}_{S})}{\bigl(d(\mathcal{\tilde{D}},\mathcal{D}_{S})+d(\mathcal{\tilde{D}},\mathcal{D}_{T})\bigr)\tau}\Bigr)},
\end{equation}
where $\tau$ is a temperature constant set at $0.05$. $q$ should be small initially, and the moving average formula is applied to adjust $\lambda_2$ as follows:
\begin{equation}
    \label{eq:lambda_update}
    \lambda_{2}^{n}=\frac{n(1-q)}{N}+q\lambda_{2}^{n-1},
\end{equation}
where $N$ is the total number of epochs. The uniform distribution $U(.)$ along with the random perturbation is introduced to stabilize the training process. 
\begin{equation}
    \tilde{\lambda}_{2}^{n}=\operatorname{Clamp}\bigl(U(\lambda_{2}^{n}-\sigma,\lambda_{2}^{n}+\sigma),\min=0.0,\max=1.0\bigr),
\end{equation}
where $\sigma$ is a local perturbation fixed at $0.2$. $\tilde{\lambda}_{2}^{n}\in[0,1]$ is sampled from the uniform distribution and clamped at the range of $[0,1]$.

Once created, the intermediate domain $\mathcal{\tilde{D}}$ is learned via the episodic-based meta-learning procedure. 
\begin{equation}
    \label{eq:loss_intermediate}
    \mathcal{L}_{\text{inter}}=\mathbb{E}_{(x,y)\backsim\mathcal{\tilde{D}}}\Bigl[l\Bigl(f_{\phi}\bigl(g_{\psi}(\tilde{x})\bigr),\tilde{y}\Bigr)+l\Bigl(f_{\phi}\bigl(\tilde{g}_{\psi}(x)\bigr),\tilde{y}\Bigr)\Bigr].
\end{equation}
This dampens the domain gap by learning an auxiliary domain, the intermediate domain, by controlling the mixup ratio. 

Overall, the episodic-based meta-learning strategy is implemented for all steps. First, we train the network on the source domain with $\mathcal{L}_{S}$ loss. Second, we train the network on the support set of the target domain by the $\mathcal{L}_{\hat{T}}$ loss. This step includes pseudo-labeling, label smoothing, and refining the samples by selecting those with the highest confidence scores. Finally, we train the network on the intermediate domain by optimizing the $\mathcal{L}_{\text{inter}}$. Moreover, we apply label smoothing at every inference step.

\begin{figure*}[t!]
    \centering
    \includegraphics[width=0.8\linewidth]{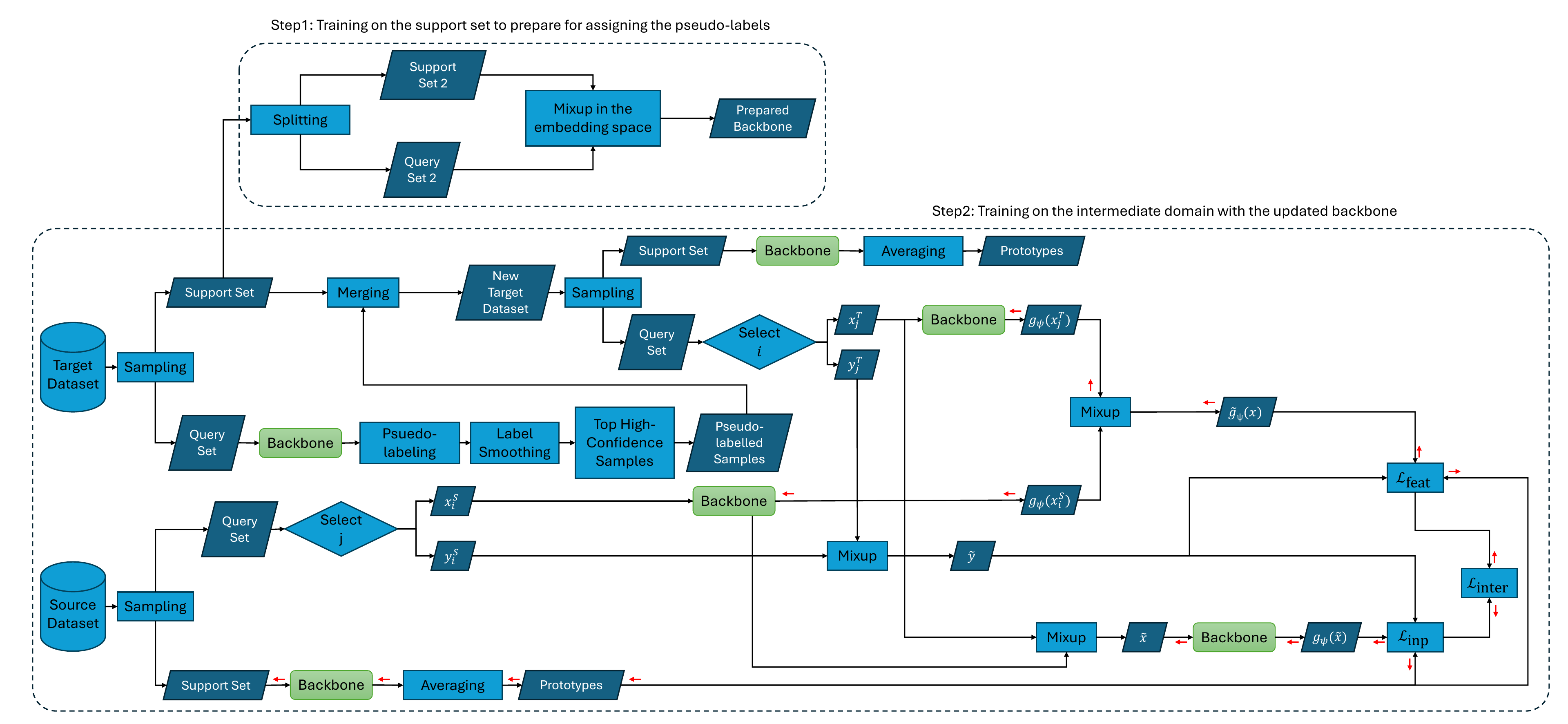}
    \caption{The flowchart of the single iteration of intermediate domain training. The red arrows indicate the backpropagated gradients.}
    \label{fig:Flowchart_Intermediate}
\end{figure*}

Fig.~\ref{fig:Flowchart_Intermediate} depicts the intermediate domain training phase. It is worth noting that when we perform mixup in the embedding space, the gradients are backpropagated to the original pair of samples, whereas when we perform mixup at the input level, the gradients flow to the single mixed sample in the middle of the two drawn samples.

\subsection{Algorithm}
The pseudo-code of the MIFOMO is offered in Algorithm \ref{alg:MIFOMO}. The training procedure starts with the source domain using both original samples and mixed samples. It proceeds to the intermediate domain afterward, where the pseudo-labeling phase is carried out. That is, target-domain training is executed using the support set of the target domain. This step is required since the source and the target domains have disjoint label spaces. Specifically, the support and query subsets are sampled from the original support set to train the model in the supervised mode. This allows the generation of initial pseudo-labels further refined by the label smoothing approach. We select the top-k highest confidence scores of the query samples to induce the final pseudo-labels. Once completing the pseudo-labeling step, the intermediate domain is constructed by mixing up the source and target domains using the adaptive mixup ratio. The model is then learned ordinarily using the supervised learning loss.

\subsection{Time-complexity Analysis}
The HyperSIGMA backbone has two ViT branches. Each branch requires the same $O(L (d n^2 + n d^2 + n d d_{\text{MLP}}))$, where $L$ is the number of layers, $n$ is the number of tokens, $d$ is the dimension of the embeddings, and $d_{\text{MLP}}$ is the dimension of the hidden layers of the of MLPs. In practice, we forward only the first 4 layers of the spectral branch for fusion and use a few linear layers and 4 negligible shallow MLPs, so the utilization is less than two branches. Moreover, the CP adds only $O(\frac{n d^2}{n_h} + n^2 d)$ per layer for $n_h$ heads. Therefore, the overall time-complexity is still $O(L (d n^2 + n d^2 + ndd d_{\text{MLP}}))$.

\section{Experiments}

\subsection{Datasets Specifications}

To evaluate the effectiveness of the proposed method, we use a widely used, publicly available CDFSL for HSI classification datasets. We use the Chikusei dataset (CH) \cite{yokoya2016airborne} for the source domain. Moreover, we use the Indian Pines (IP), Salinas (SA), and Pavia University (PU) datasets for the target domain \cite{Li2021DeepCF}. \par

The CH dataset was collected using the Headwall Hyperpec-VNIR-C sensor in Chikusei, Ibaraki, Japan, on July 29, 2014, which comprises 128 bands spanning 340-1018 nm. The image is 2517x2335 and contains 19 categories. More details are provided in \cref{tab:ch_dataset}. \par

The IP dataset was acquired by AVIRIS in 1992 over Northwest Indiana and contains 200 bands spanning 400-2500 nm, with a spatial size of 145x145 and 16 classes. The SA dataset was taken by the AVIRIS sensor in the Salinas Valley, California, USA. The image size is 512x217. The spectral image contains 224 bands, of which 204 bands are useful. The PU dataset was taken by the ROSIS sensor over Pavia, in northern Italy, in 2003. It contains 103 bands after discarding the noise bands, and covers the range 430-860 nm. The image size is 620x340 and has 9 classes. More details about the target datasets are provided in Tables \ref{tab:ch_dataset}-\ref{tab:ht_dataset} \cite{Shi2025MultiGaussianPM, Li2024CrossDomainFH}.\par

\begin{table}[htbp]
    \centering
    \caption{Class names and samples per class for Chikusei dataset}
    \label{tab:ch_dataset}
    \begin{tabular}{|l|l|l|}
    \hline
    \textbf{Class} & \textbf{Name} & \textbf{Samples} \\ \hline
        1 & Water & 2845 \\ \hline
        2 & Bare soil (school) & 2859 \\ \hline
        3 & Bare soil (park) & 286 \\ \hline
        4 & Bare soil (farmland) & 48525 \\ \hline
        5 & Natural plants & 4297 \\ \hline
        6 & Weeds in farmland & 1108 \\ \hline
        7 & Forest & 20516 \\ \hline
        8 & Grass & 6515 \\ \hline
        9 & Rice field (grown) & 13369 \\ \hline
        10 & Rice field (first stage) & 1268 \\ \hline
        11 & Row crops & 5961 \\ \hline
        12 & Plastic house & 2193 \\ \hline
        13 & Manmade (non-dark) & 1220 \\ \hline
        14 & Manmade (dark) & 7664 \\ \hline
        15 & Manmade (blue) & 431 \\ \hline
        16 & Manmade (red) & 222 \\ \hline
        17 & Manmade grass & 1040 \\ \hline
        18 & Asphalt & 801 \\ \hline
        19 & Paved ground & 145 \\ \hline
        \multicolumn{2}{|l|}{\textbf{Total}} & \textbf{77592} \\ \hline
    \end{tabular}
\end{table}
\begin{table}[htbp]
    \centering
    \caption{Class names and samples per class for the Indian Pines dataset}
    \label{tab:ip_dataset}
    \begin{tabular}{|l|l|l|}
    \hline
    \textbf{Class} & \textbf{Name} & \textbf{Samples} \\ \hline
        1 & Alfalfa & 46 \\ \hline
        2 & Corn-notil & 1428 \\ \hline
        3 & Corn-mintill & 830 \\ \hline
        4 & Corn & 237 \\ \hline
        5 & Grass pasture & 483 \\ \hline
        6 & Grass trees & 730 \\ \hline
        7 & Grass pasture mowed & 28 \\ \hline
        8 & Hay windrowed & 478 \\ \hline
        9 & Oats & 20 \\ \hline
        10 & Soybean notill & 972 \\ \hline
        11 & Soybean mintill & 2455 \\ \hline
        12 & Soybean clean & 593 \\ \hline
        13 & Wheat & 205 \\ \hline
        14 & Woods & 1265 \\ \hline
        15 & Buildings Grass Trees Drives & 386 \\ \hline
        16 & Stone Steel Towers & 93 \\ \hline
        \multicolumn{2}{|l|}{\textbf{Total}} & \textbf{10249} \\ \hline
    \end{tabular}
\end{table}
\begin{table}[htbp]
\centering
\caption{Class names and samples per class for the Salinas dataset}
\label{tab:sa_dataset}
    \begin{tabular}{|l|l|l|}
    \hline
    \textbf{Class} & \textbf{Name} & \textbf{Samples} \\ \hline
        1 & Brocoli green weeds 1 & 2009 \\ \hline
        2 & Brocoli green weeds 2 & 3726 \\ \hline
        3 & Fallow & 1976 \\ \hline
        4 & Fallow rough plow & 1394 \\ \hline
        5 & Fallow smooth & 2678 \\ \hline
        6 & Stubble & 3959 \\ \hline
        7 & Celery & 3579 \\ \hline
        8 & Grapes untrained & 11271 \\ \hline
        9 & Soil vinyard develop & 6203 \\ \hline
        10 & Corn senesced green weeds & 3278 \\ \hline
        11 & Lettuce romaine 4wk & 1068 \\ \hline
        12 & Lettuce romaine 5wk & 1927 \\ \hline
        13 & Lettuce romaine 6wk & 916 \\ \hline
        14 & Lettuce romaine 7wk & 1070 \\ \hline
        15 & Vinyard untrained & 7268 \\ \hline
        16 & Vinyard untrained trellis & 1807 \\ \hline
        \multicolumn{2}{|l|}{\textbf{Total}} & \textbf{54129} \\ \hline
    \end{tabular}
\end{table}
\begin{table}[htbp]
\centering
\caption{Class names and samples per class for the Pavia University dataset}
\label{tab:pu_dataset}
    \begin{tabular}{|l|l|l|}
    \hline
    \textbf{Class} & \textbf{Name} & \textbf{Samples} \\ \hline
        1 & Asphalt & 6631 \\ \hline
        2 & Meadows & 18649 \\ \hline
        3 & Gravel & 2099 \\ \hline
        4 & Trees & 3064 \\ \hline
        5 & Painted metal & 1345 \\ \hline
        6 & Bare Soil & 5029 \\ \hline
        7 & Bitumen & 1330 \\ \hline
        8 & Self-blocking & 3682 \\ \hline
        9 & Shadows & 947 \\ \hline
        \multicolumn{2}{|l|}{\textbf{Total}} & \textbf{42776} \\ \hline
    \end{tabular}
\end{table}
\begin{table}[htbp]
    \centering
    \caption{Number of training and testing set samples per class in the Houston dataset}
    \label{tab:ht_dataset}
    \begin{tabular}{|l|l|l|l|}
    \hline
    \textbf{Class} & \textbf{Name} & \textbf{Training Samples} & \textbf{Testing Samples} \\ \hline
        1 & Healthy grass & 198 & 1053 \\ \hline
        2 & Stressed grass & 190 & 1064 \\ \hline
        3 & Synthetic grass & 192 & 505 \\ \hline
        4 & Trees & 188 & 1056 \\ \hline
        5 & Soil & 186 & 1056 \\ \hline
        6 & Water & 182 & 143 \\ \hline
        7 & Residential & 196 & 1072 \\ \hline
        8 & Commercial & 191 & 1053 \\ \hline
        9 & Road & 193 & 1059 \\ \hline
        10 & Highway & 191 & 1036 \\ \hline
        11 & Railway & 181 & 1054 \\ \hline
        12 & Parking Lot 1 & 192 & 1041 \\ \hline
        13 & Parking Lot 2 & 184 & 285 \\ \hline
        14 & Tennis Court & 181 & 247 \\ \hline
        15 & Running Track & 187 & 473 \\ \hline
        \multicolumn{2}{|l|}{\textbf{Total}} & \textbf{2832} & \textbf{12197} \\ \hline
    \end{tabular}
\end{table}

\subsection{Baseline methods}

To evaluate the effectiveness of our method, we comprehensively compare it against the MGPDO~\cite{Shi2025MultiGaussianPM}, CFSL-KT~\cite{Huang2023HyperspectralIC}, CDLA~\cite{Yu2025FewshotLF}, CF-Trans~\cite{Zhu2025FromIT}, CMTL~\cite{Cheng2023CausalML}, GMTN~\cite{Wang2023GraphMT}, DCFSL~\cite{Li2021DeepCF}, CMFSL~\cite{Xi2022FewShotLW}, HFSL~\cite{Wang2022HeterogeneousFL}, CNN-IT~\cite{Huang2020HyperspectralIC}, Gia-CFSL~\cite{Zhang2022GraphIA}, SSRN~\cite{Zhong2018SpectralSpatialRN}, MCNN~\cite{zheng2020hyperspectral}, SPRN~\cite{zhang2021spectral}, FDFSL~\cite{Qin2024CrossDomainFL}, ADAFSL~\cite{Ye2023AdaptiveDF}, RPCL-FSL~\cite{liu2023category}, CASCL~\cite{Li2024CrossDomainFH}, GLGAT-CFSL~\cite{Ding2024GLGATCFSLGG}, and DA-CFSL~\cite{Qin2025DualalignmentCF}.

\subsection{Implementation Details}

The experiments are conducted on a workstation with an NVIDIA GeForce RTX 4090 GPU and Intel Core i9-14900K. We performed the experiments with 20 episodes (random seeds). We utilize the ViT-B model from HyperSIGMA as the backbone with two ViT branches pre-trained with masked autoencoder (MAE)~\cite{He2021MaskedAA} on the HyperGlobal-450K dataset~\cite{Wang2024HyperSIGMAHI}. Moreover, we use Principal Component Analysis (PCA) to decrease the number of bands to 50 for all datasets as a preprocessing step.

\begin{table*}[htbp]
\centering
\caption{Class-specific classification accuracy (\%), OA (\%), AA (\%), and KC (\%) on the Indian pines dataset (Five labeled samples per class)}
\label{tab:ip_results}
\resizebox{\textwidth}{!}{

    \begin{tabular}{|c|c|c|c|c|c|c|c|c|c|c|c|c|c|c|c|c|c|c|c|}
    \hline
    Class & 1 & 2 & 3 & 4 & 5 & 6 & 7 & 8 & 9 & 10 & 11 & 12 & 13 & 14 & 15 & 16 & OA (\%) & AA (\%) & KC (\%) \\ 
    \hline
    SSRN~\cite{Zhong2018SpectralSpatialRN} & 99.01 & 42.91 & 58.84 & 81.49 & 78.63 & 95.75 & 99.55 & 76.91 & 100 & 60.51 & 51.14 & 53.59 & 99.29 & 87.98 & 75.21 & 97.71 & \val{65.72}{3.77} & \val{79.22}{1.62} & \val{61.72}{3.90} \\ 
    MCNN~\cite{zheng2020hyperspectral} & 93.24 & 46.14 & 46.24 & 76.57 & 66.39 & 87.15 & 97.27 & 78.14 & 100 & 65.32 & 62.34 & 53.17 & 97.14 & 87.24 & 75.96 & 97.58 & \val{66.17}{3.21} & \val{80.32}{2.83} & \val{62.57}{3.67} \\ 
    SPRN~\cite{zhang2021spectral} & 98.94 & 54.76 & 51.14 & 83.78 & 73.53 & 90.91 & 99.65 & 84.13 & 100 & 67.95 & 65.22 & 49.22 & 97.58 & 90.22 & 80.18 & 96.51 & \val{67.32}{2.08} & \val{81.28}{1.12} & \val{63.32}{2.93} \\ 
    DCFSL~\cite{Li2021DeepCF} & 97.8 & 45.14 & 46.88 & 75.60 & 67.74 & 86.50 & 99.57 & 76.03 & 100 & 57.08 & 61.45 & 48.69 & 95.35 & 89.48 & 67.72 & 96.82 & \val{64.89}{3.50} & \val{75.75}{2.44} & \val{60.36}{3.96} \\ 
    Gia-CFSL~\cite{Zhang2022GraphIA} & 93.41 & 47.98 & 49.35 & 76.51 & 66.65 & 84.14 & 99.57 & 83.76 & 100 & 58.23 & 61.32 & 45.83 & 96.45 & 87.83 & 70.21 & 96.02 & \val{65.44}{2.23} & \val{76.08}{1.48} & \val{61.02}{2.53} \\ 
    FDFSL~\cite{Qin2024CrossDomainFL} & 96.59 & 55.62 & 64.05 & 97.07 & 75.17 & 83.52 & 99.57 & 83.11 & 100 & 66.10 & 64.51 & 59.29 & 98.70 & 88.60 & 80.21 & 98.18 & \val{71.11}{2.19} & \val{81.27}{1.84} & \val{67.55}{2.44} \\ 
    ADAFSL~\cite{Ye2023AdaptiveDF} & 100 & 51.03 & 64.10 & 81.21 & 79.75 & 92.65 & 100 & 92.30 & 100 & 59.54 & 69.93 & 59.93 & 97.20 & 87.86 & 71.50 & 97.73 & \val{71.91}{5.15} & \val{81.54}{1.49} & \val{68.23}{3.39} \\ 
    RPCL-FSL~\cite{liu2023category} & 98.54 & 62.58 & 57.09 & 93.23 & 78.16 & 90.06 & 100 & 90.51 & 100 & 73.80 & 69.56 & 64.40 & 98.50 & 90.55 & 80.63 & 97.05 & \val{75.11}{1.97} & \val{84.04}{1.10} & \val{71.94}{2.17} \\ 
    CFSL-KT~\cite{Huang2023HyperspectralIC} & 98.57 & 44.86 & 72.30 & 88.45 & 74.94 & 96.80 & 100.0 & 99.24 & 100.0 & 78.82 & 73.99 & 61.87 & 100.0 & 91.71 & 83.66 & 99.77 & \val{76.29}{1.26} & \val{85.36}{1.21} & \val{73.41}{1.39} \\
    GLGAT-CFSL~\cite{Ding2024GLGATCFSLGG} & 100 & 78.83 & 91.93 & 95.79 & 96.99 & 95.03 & 99.74 & 98.91 & 99.67 & 79.82 & 80.96 & 70.03 & 98.75 & 97.84 & 68.93 & 100 & 78.5 & 82.96 & 79.98 \\ 
    CDLA~\cite{Yu2025FewshotLF} & 99.27 & 68.01 & 73.56 & 92.24 & 80.77 & 93.23 & 100.0 & 95.18 & 100.0 & 71.61 & 70.99 & 71.43 & 97.90 & 92.49 & 91.39 & 98.07 & \val{78.94}{3.88} & \val{87.26}{1.59} & \val{76.26}{4.23} \\ 
    DA-CFSL~\cite{Qin2025DualalignmentCF} & 99.27 & 73.23 & 73.78 & 91.25 & 82.34 & 91.75 & 100 & 94.29 & 100 & 71.32 & 70.87 & 74.78 & 97.9 & 90.71 & 84.78 & 98.75 & \val{79.26}{3.45} & \val{87.19}{1.29} & \val{76.62}{3.73} \\ 
    CASCL~\cite{Li2024CrossDomainFH} & 97.56 & 73.11 & 62.85 & 95.32 & 77.86 & 94.86 & 100 & 95.32 & 100 & 76.25 & 78.14 & 63.20 & 98.33 & 89.13 & 86.75 & 97.78 & \val{79.73}{1.48} & \val{86.34}{1.73} & \val{76.53}{1.86} \\ 
    MGPDO~\cite{Shi2025MultiGaussianPM} & 100 & 71.98 & 63.56 & 90.37 & 83.47 & 96.37 & 100 & 99.93 & 100 & 75.32 & 80.98 & 59.41 & 99.33 & 92.67 & 91.69 & 95.45 & \val{81.32}{1.62} & \val{87.53}{0.31} & \val{78.75}{1.73} \\ 
    MIFOMO & 88.33 & 83.02 & 93.77 & 73.23 & 81.75 & 77.69 & 91.76 & 97.77 & 83.69 & 94.05 & 96.44 & 87.32 & 87.63 & 89.46 & 91.55 & 98.39 & \val{95.44}{3.19} & \val{88.74}{3.93} & \val{94.82}{3.62} \\
    \hline
    \end{tabular}
}
\end{table*}
\begin{table*}[htbp]
    \centering
    \caption{Class-specific classification accuracy (\%), OA (\%), AA (\%), and KC (\%) on the Pavia University dataset (Five labeled samples per class)}
    \label{tab:pu_results}
    \resizebox{\textwidth}{!}{
    
    \begin{tabular}{|c|c|c|c|c|c|c|c|c|c|c|c|c|}
    \hline
    Class & 1 & 2 & 3 & 4 & 5 & 6 & 7 & 8 & 9 & OA (\%) $\downarrow$ & AA (\%) & KC (\%) \\ 
    \hline
    SSRN~\cite{Zhong2018SpectralSpatialRN} & 91.21 & 93.94 & 55.30 & 95.21 & 99.32 & 45.06 & 77.42 & 73.58 & 95.44 & \val{76.78}{2.01} & \val{80.08}{2.54} & \val{69.32}{2.52} \\
    MCNN~\cite{zheng2020hyperspectral} & 81.23 & 86.31 & 57.62 & 94.23 & 99.13 & 72.82 & 77.41 & 69.63 & 95.57 & \val{77.23}{2.25} & \val{80.89}{2.74} & \val{70.25}{2.62} \\
    SPRN~\cite{zhang2021spectral} & 82.63 & 86.85 & 58.88 & 96.13 & 99.03 & 75.97 & 78.64 & 67.61 & 96.81 & \val{79.52}{1.74} & \val{82.29}{1.49} & \val{73.64}{2.01} \\
    DCFSL~\cite{Li2021DeepCF} & 80.37 & 85.49 & 63.74 & 94.27 & 99.51 & 76.30 & 79.42 & 59.94 & 98.67 & \val{81.52}{2.72} & \val{81.97}{1.23} & \val{76.10}{2.71} \\
    Gia-CFSL~\cite{Zhang2022GraphIA} & 88.12 & 85.95 & 66.44 & 92.78 & 99.18 & 75.03 & 74.58 & 72.42 & 94.67 & \val{83.60}{2.29} & \val{83.21}{1.86} & \val{78.67}{2.73} \\
    FDFSL~\cite{Qin2024CrossDomainFL} & 86.07 & 83.09 & 65.16 & 89.29 & 99.40 & 75.11 & 88.66 & 85.19 & 78.34 & \val{83.31}{1.89} & \val{85.19}{2.23} & \val{78.34}{2.16} \\
    ADAFSL~\cite{Ye2023AdaptiveDF} & 87.81 & 86.92 & 59.96 & 88.58 & 99.36 & 78.47 & 86.20 & 82.26 & 99.92 & \val{85.12}{1.99} & \val{85.50}{1.78} & \val{80.59}{2.41} \\
    RPCL-FSL~\cite{liu2023category} & 87.57 & 87.53 & 67.62 & 94.44 & 99.02 & 59.45 & 85.88 & 86.89 & 99.36 & \val{86.44}{1.97} & \val{85.86}{2.19} & \val{82.05}{1.73} \\
    CASCL~\cite{Li2024CrossDomainFH} & 89.50 & 86.44 & 75.64 & 93.34 & 99.48 & 85.21 & 91.18 & 86.43 & 99.52 & \val{89.75}{1.34} & \val{91.32}{1.86} & \val{85.52}{1.96} \\
    GLGAT-CFSL~\cite{Ding2024GLGATCFSLGG} & 95.85 & 96.95 & 85.09 & 91.92 & 98.8 & 92.57 & 92.28 & 90.95 & 98.98 & 88.17 & 91.71 & 88.35 \\
    CDLA~\cite{Yu2025FewshotLF} & 88.48 & 84.38 & 80.92 & 90.89 & 98.84 & 86.37 & 97.77 & 88.9 & 97.27 & \val{87.09}{4.35} & \val{90.42}{1.87} & \val{83.40}{5.13} \\
    DA-CFSL~\cite{Qin2025DualalignmentCF} & 92.95 & 87.13 & 83.71 & 88.78 & 99.6 & 84.44 & 97.65 & 90.47 & 96.44 & \val{88.88}{2.70} & \val{91.24}{2.57} & \val{85.57}{3.35} \\
    CFSL-KT~\cite{Huang2023HyperspectralIC} & 98.79 & 86.36 & 86.71 & 93.95 & 99.25 & 96.92 & 99.92 & 73.78 & 94.42 & \val{89.92}{2.50} & \val{92.22}{1.93} & \val{87.03}{3.11} \\
    MGPDO~\cite{Shi2025MultiGaussianPM} & 92.09 & 92.50 & 89.85 & 96.14 & 99.46 & 98.71 & 98.91 & 90.75 & 99.61 & \val{91.08}{1.07} & \val{94.66}{0.91} & \val{88.50}{1.37} \\
    MIFOMO & 96.33 & 99.33 & 98.81 & 95.96 & 98.59 & 97.75 & 99.67 & 98.58 & 97.70 & \val{97.76}{2.90} & \val{98.08}{2.00} & \val{97.07}{3.75} \\
    \hline
    \end{tabular}
    }
\end{table*}
\begin{table*}[htbp]
    \centering
    \caption{Class-specific classification accuracy (\%), OA (\%), AA (\%), and KC (\%) on the Salinas dataset (Five labeled samples per class)}
    \label{tab:sa_results}
    \resizebox{\textwidth}{!}{

    \begin{tabular}{|c|c|c|c|c|c|c|c|c|c|c|c|c|c|c|c|c|c|c|c|}
    \hline
    Class & 1 & 2 & 3 & 4 & 5 & 6 & 7 & 8 & 9 & 10 & 11 & 12 & 13 & 14 & 15 & 16 & OA (\%) & AA (\%) & KC (\%) \\ 
    \hline
    SSRN~\cite{Zhong2018SpectralSpatialRN} & 98.53 & 99.67 & 96.87 & 99.31 & 95.79 & 99.66 & 99.90 & 66.17 & 99.79 & 90.73 & 97.34 & 99.88 & 99.01 & 98.13 & 81.47 & 94.43 & \val{89.13}{2.59} & \val{94.79}{0.55} & \val{87.95}{2.80} \\ 
    MCNN~\cite{zheng2020hyperspectral} & 98.41 & 99.76 & 91.96 & 99.25 & 92.70 & 99.52 & 98.58 & 74.57 & 99.59 & 83.42 & 96.61 & 98.93 & 98.30 & 98.24 & 76.38 & 92.22 & \val{89.34}{2.19} & \val{94.04}{1.14} & \val{88.17}{2.41} \\ 
    SPRN~\cite{zhang2021spectral} & 99.13 & 99.69 & 97.01 & 99.36 & 94.58 & 99.47 & 99.53 & 78.74 & 99.13 & 84.98 & 96.99 & 99.13 & 98.34 & 98.36 & 80.54 & 91.03 & \val{90.05}{1.53} & \val{94.11}{0.96} & \val{88.63}{1.91} \\ 
    DCFSL~\cite{Li2021DeepCF} & 99.45 & 98.82 & 97.31 & 99.26 & 91.52 & 99.84 & 99.69 & 79.48 & 99.32 & 86.78 & 97.01 & 99.17 & 97.85 & 98.43 & 75.73 & 92.75 & \val{90.51}{0.99} & \val{94.52}{0.66} & \val{89.45}{1.09} \\ 
    Gia-CFSL~\cite{Zhang2022GraphIA} & 98.09 & 99.41 & 95.46 & 98.52 & 86.55 & 99.25 & 99.28 & 81.31 & 97.76 & 83.73 & 93.56 & 99.19 & 94.27 & 98.46 & 74.51 & 90.32 & \val{89.75}{0.80} & \val{93.10}{1.19} & \val{88.60}{0.90} \\ 
    FDFSL~\cite{Qin2024CrossDomainFL} & 98.96 & 98.78 & 80.10 & 99.32 & 90.47 & 98.41 & 99.76 & 78.85 & 97.98 & 85.40 & 95.51 & 99.64 & 98.29 & 99.00 & 82.53 & 92.04 & \val{90.51}{1.17} & \val{93.94}{1.26} & \val{89.45}{1.30} \\ 
    CF-Trans~\cite{Zhu2025FromIT} & 99.08 & 99.81 & 95.38 & 99.44 & 92.32 & 99.13 & 99.47 & 80.06 & 98.97 & 87.31 & 98.53 & 99.76 & 98.78 & 98.82 & 79.18 & 91,27 & \val{91.03}{1.25} & \val{94.74}{0.69} & \val{90.03}{1.38} \\
    ADAFSL~\cite{Ye2023AdaptiveDF} & 99.19 & 99.68 & 97.10 & 99.57 & 95.74 & 99.26 & 99.30 & 77.67 & 99.66 & 84.64 & 99.08 & 99.92 & 99.09 & 99.44 & 84.40 & 92.62 & \val{91.43}{0.47} & \val{95.28}{0.58} & \val{90.49}{0.52} \\ 
    RPCL-FSL~\cite{liu2023category} & 98.95 & 99.09 & 96.09 & 99.09 & 90.83 & 98.76 & 99.17 & 82.24 & 99.94 & 90.00 & 99.13 & 98.36 & 97.09 & 97.50 & 84.59 & 93.14 & \val{92.33}{0.78} & \val{95.22}{0.71} & \val{91.47}{0.83} \\ 
    CASCL~\cite{Li2024CrossDomainFH} & 99.45 & 99.62 & 99.84 & 99.49 & 95.94 & 99.82 & 99.91 & 84.52 & 99.71 & 92.69 & 98.30 & 99.81 & 99.66 & 98.24 & 83.91 & 94.23 & \val{93.64}{0.63} & \val{95.89}{0.42} & \val{93.07}{0.69} \\ 
    CDLA~\cite{Yu2025FewshotLF} & 99.87 & 99.77 & 92.37 & 99.67 & 96.0 & 99.61 & 98.66 & 83.55 & 99.98 & 89.95 & 99.70 & 99.71 & 99.48 & 99.43 & 82.05 & 91.47 & \val{92.62}{1.38} & \val{95.72}{0.95} & \val{91.79}{1.53} \\
    CFSL-KT~\cite{Huang2023HyperspectralIC} & 97.51 & 99.38 & 99.85 & 99.64 & 98.10 & 99.44 & 100.0 & 84.67 & 99.98 & 92.80 & 99.81 & 100.0 & 99.67 & 99.72 & 93.31 & 100.0 & \val{95.13}{0.93} & \val{97.69}{0.52} & \val{94.60}{1.03} \\
    MGPDO~\cite{Shi2025MultiGaussianPM} & 100 & 100 & 99.81 & 99.81 & 97.27 & 99.75 & 99.63 & 86.81 & 100 & 93.39 & 99.92 & 99.75 & 99.58 & 99.64 & 85.98 & 94.23 & \val{94.19}{0.94} & \val{96.99}{0.40} & \val{93.54}{1.05} \\ 
    MIFOMO & 99.51 & 94.29 & 93.11 & 99.28 & 99.40 & 99.50 & 99.20 & 99.46 & 99.11 & 99.51 & 99.53 & 99.70 & 99.44 & 99.39 & 99.06 & 97.02 & \val{96.57}{6.13} & \val{98.53}{1.86} & \val{96.23}{6.73} \\
    \hline
    \end{tabular}
    }
\end{table*}
\begin{table*}[htbp]
    \centering
    \caption{Class-specific classification accuracy (\%), OA (\%), AA (\%), and KC (\%) on the Houston dataset (Five labeled samples per class)}
    \label{tab:hs_results}
    \resizebox{\textwidth}{!}{

    \begin{tabular}{|c|c|c|c|c|c|c|c|c|c|c|c|c|c|c|c|c|c|c|}
    \hline
    Class & 1 & 2 & 3 & 4 & 5 & 6 & 7 & 8 & 9 & 10 & 11 & 12 & 13 & 14 & 15 & OA (\%) $\downarrow$ & AA (\%) & KC (\%) \\ 
    \hline 
    SSRN~\cite{Zhong2018SpectralSpatialRN} & 73.00 & 83.98 & 92.83 & 80.95 & 94.44 & 77.91 & 67.28 & 25.47 & 81.09 & 52.55 & 49.67 & 25.03 & 81.47 & 99.27 & 98.76 & \val{68.23}{4.21} & \val{72.25}{4.16} & \val{65.44}{4.75} \\ 
    Gia-CFSL~\cite{Zhang2022GraphIA} & 88.80 & 82.19 & 94.78 & 89.90 & 94.32 & 73.72 & 63.33 & 42.66 & 59.17 & 63.98 & 62.50 & 72.52 & 88.47 & 93.31 & 92.41 & \val{75.05}{1.62} & \val{77.47}{1.69} & \val{73.05}{1.74} \\ 
    CNN-IT~\cite{Huang2020HyperspectralIC} & 82.41 & 78.80 & 98.86 & 77.14 & 93.49 & 90.81 & 74.04 & 53.83 & 67.76 & 93.46 & 84.76 & 68.84 & 53.99 & 99.43 & 99.79 & \val{79.58}{2.85} & \val{81.16}{2.62} & \val{77.92}{3.08} \\ 
    HFSL~\cite{Wang2022HeterogeneousFL} & 89.61 & 79.43 & 99.65 & 90.72 & 91.09 & 90.97 & 89.93 & 44.02 & 68.67 & 68.66 & 68.76 & 62.84 & 68.38 & 96.47 & 96.77 & \val{77.94}{2.38} & \val{80.20}{2.27} & \val{76.16}{2.56} \\ 
    CMFSL~\cite{Xi2022FewShotLW} & 89.74 & 87.47 & 95.78 & 92.28 & 97.38 & 79.93 & 73.76 & 48.04 & 70.12 & 69.35 & 66.73 & 70.94 & 95.60 & 96.00 & 97.72 & \val{79.63}{2.60} & \val{82.06}{1.40} & \val{77.99}{2.21} \\ 
    DCFSL~\cite{Li2021DeepCF} & 89.13 & 83.25 & 97.12 & 92.11 & 97.09 & 80.72 & 70.13 & 41.92 & 64.15 & 63.23 & 62.15 & 74.42 & 88.04 & 91.58 & 97.66 & \val{77.00}{1.71} & \val{79.51}{1.62} & \val{75.14}{1.86} \\ 
    GMTN~\cite{Wang2023GraphMT} & 96.14 & 82.87 & 99.42 & 86.12 & 92.40 & 84.69 & 80.60 & 45.84 & 73.13 & 55.07 & 62.28 & 60.67 & 86.85 & 96.22 & 99.39 & 77.20 & 80.11 & 75.38 \\ 
    CMTL~\cite{Cheng2023CausalML} & 97.43 & 63.01 & 100.0 & 93.46 & 99.19 & 85.63 & 70.23 & 52.22 & 63.03 & 80.03 & 81.06 & 93.08 & 81.68 & 98.34 & 89.31 & 81.38 & 83.18 & 79.88 \\ 
    CF-Trans~\cite{Zhu2025FromIT} & 98.66 & 84.07 & 96.05 & 91.71 & 94.77 & 88.02 & 84.64 & 40.68 & 79.95 & 67.29 & 72.52 & 78.77 & 96.12 & 90.23 & 99.34 & \val{82.02}{0.53} & \val{84.19}{0.78} & \val{80.57}{0.58} \\ 
    CFSL-KT~\cite{Huang2023HyperspectralIC} & 85.12 & 84.96 & 99.86 & 83.31 & 98.63 & 94.06 & 81.73 & 56.29 & 73.52 & 91.79 & 86.99 & 66.10 & 71.49 & 100.0 & 99.70 & \val{82.63}{1.68} & \val{84.35}{1.70} & \val{81.23}{1.80} \\ 
    MIFOMO & 98.53 & 97.52 & 93.05 & 97.84 & 96.11 & 92.59 & 96.82 & 97.16 & 97.31 & 91.84 & 98.39 & 96.93 & 97.29 & 89.54 & 99.68 & \valb{95.86}{3.01} & \valb{96.04}{2.25} & \valb{95.51}{3.26} \\
    \hline
    \end{tabular}
    }
\end{table*}

\begin{figure}[htbp]
    \centering
    \subfloat[Ground Truth\label{fig:GT,IP}]{
        \includegraphics[width=0.45\linewidth]{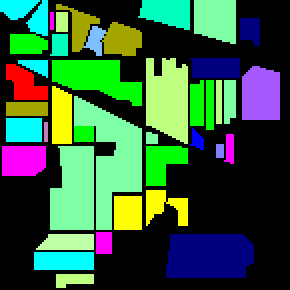}
    }
    \hfill
    \subfloat[MIFOMO\label{fig:MIFOMO,IP}]{
        \includegraphics[width=0.45\linewidth]{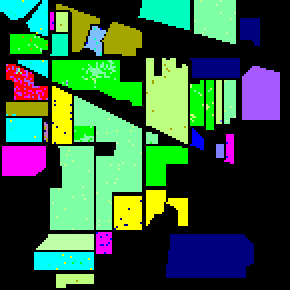}
    }\hfill

    \caption{Classification maps for the Indian Pines dataset.}  
    \label{fig:classification_map,IP}
\end{figure}

\begin{figure}[htbp]
    \centering
    \subfloat[Ground Truth\label{fig:GT,Salinas}]{
        \includegraphics[width=0.45\linewidth]{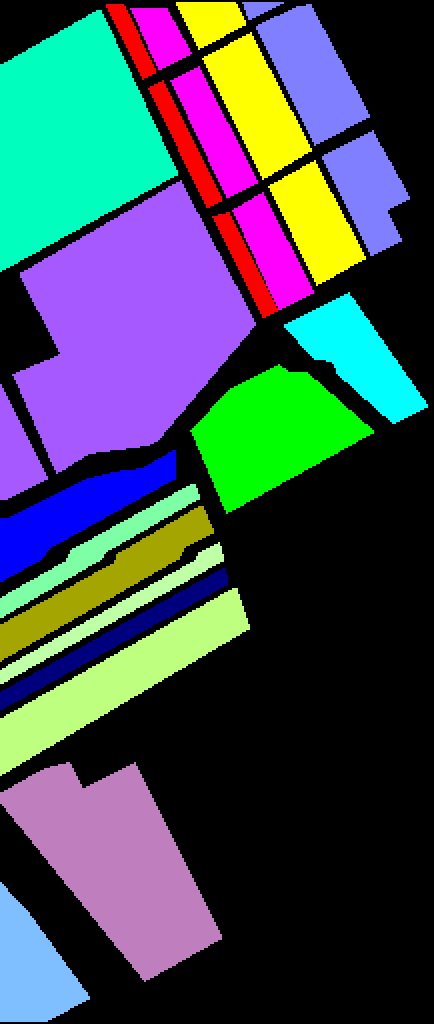}
    }
    \hfill
    \subfloat[MIFOMO\label{fig:MIFOMO,Salinas}]{
        \includegraphics[width=0.45\linewidth]{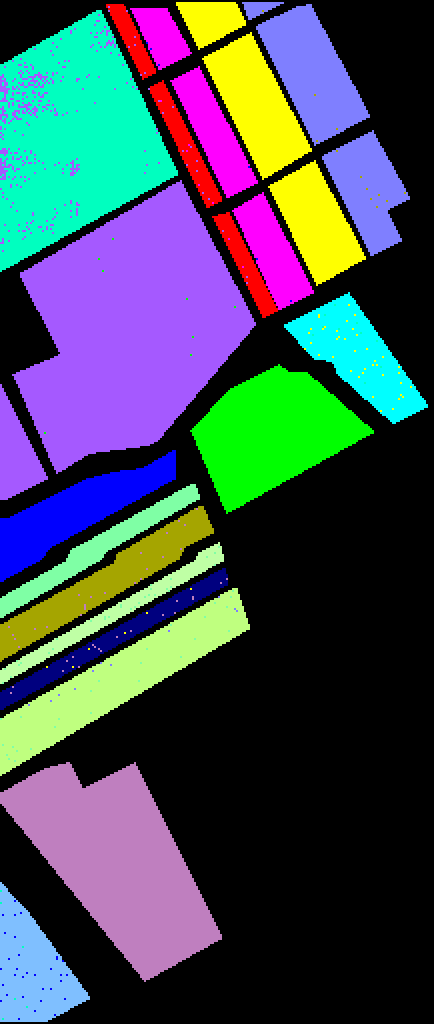}
    }\hfill

    \caption{Classification maps for the Salinas dataset.}  
    \label{fig:classification_map,salinas}
\end{figure}

\begin{figure}[htbp]
    \centering
    \subfloat[Ground Truth\label{fig:GT,PaviaU}]{
        \includegraphics[width=0.45\linewidth]{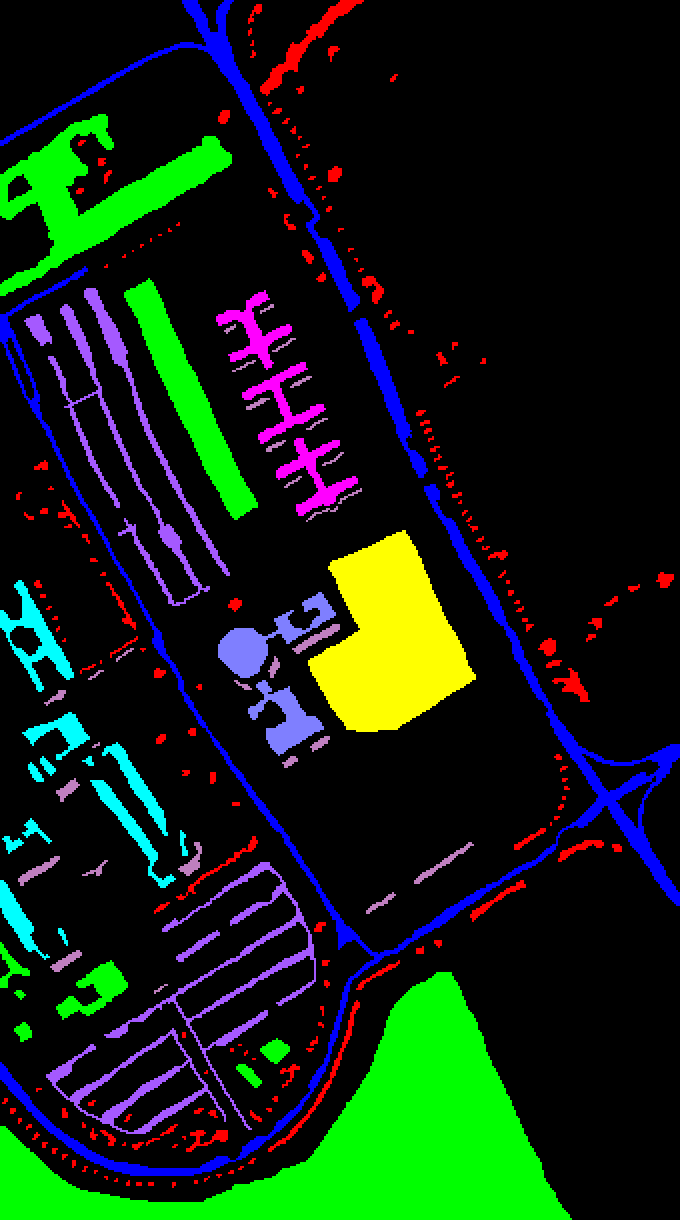}
    }
    \hfill
    \subfloat[MIFOMO\label{fig:MIFOMO,PaviaU}]{
        \includegraphics[width=0.45\linewidth]{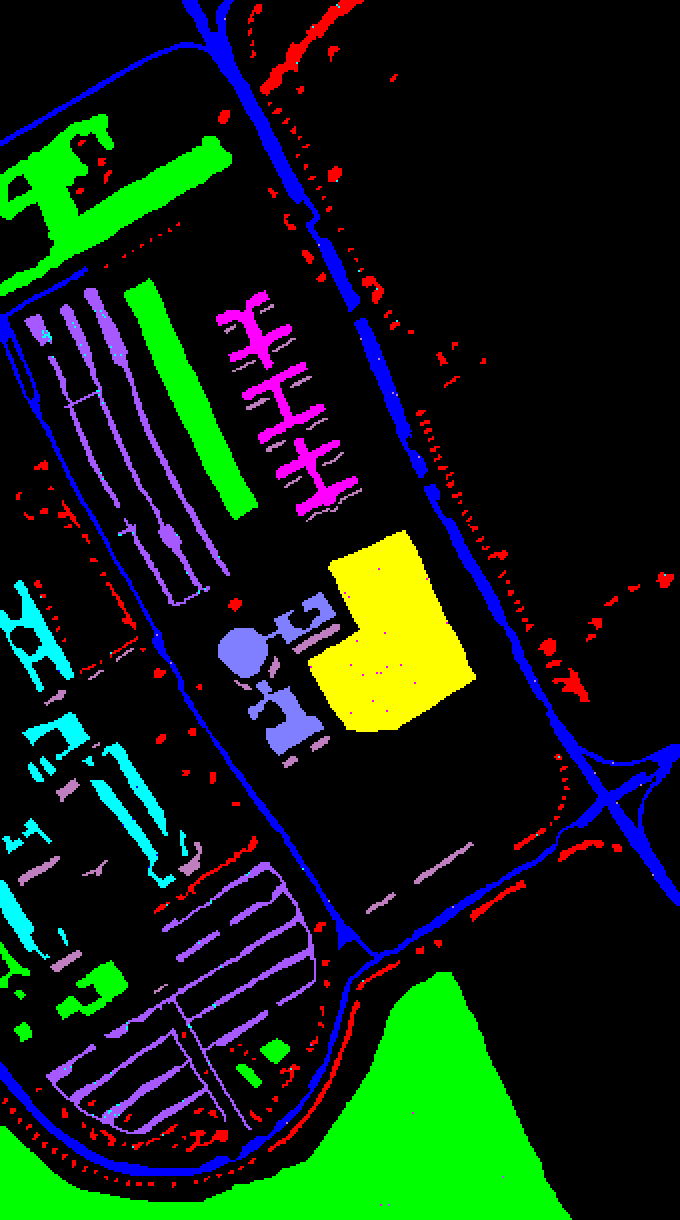}
    }\hfill

    \caption{Classification maps for the Pavia Univesity dataset.}  
    \label{fig:classification_map,pavia}
\end{figure}

\begin{figure*}[htbp]
    \centering
    \subfloat[Ground Truth\label{fig:GT,Houston}]{
        \includegraphics[width=\linewidth]{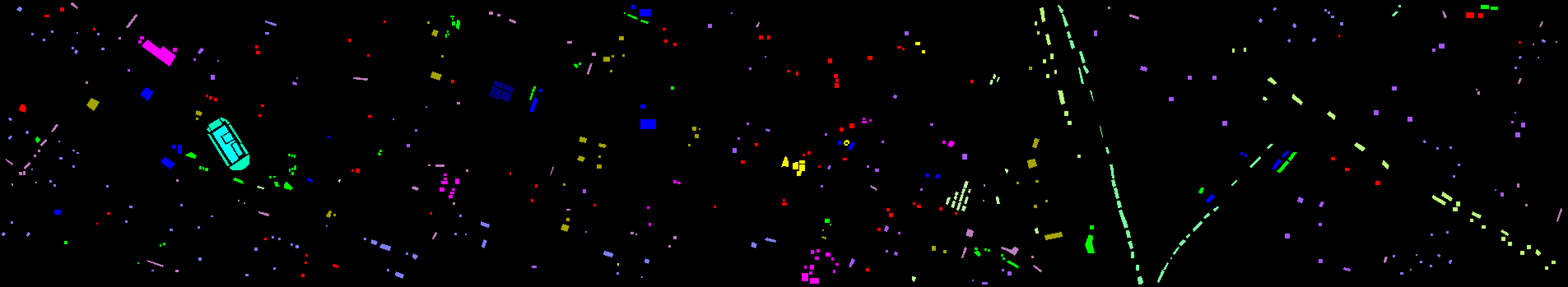}
    }

    \vspace{0.1em}

    \subfloat[MIFOMO\label{fig:MIFOMO,Houston}]{
        \includegraphics[width=\linewidth]{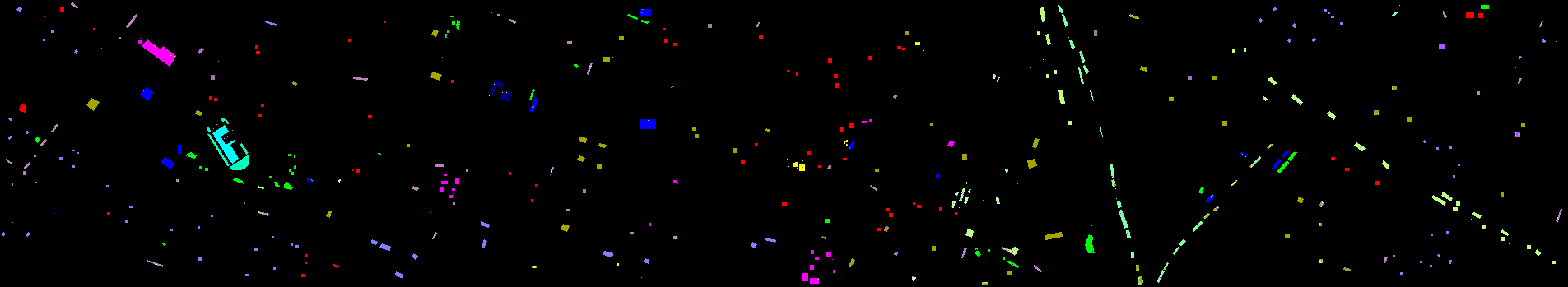}
    }\hfill

    \caption{Classification maps for the Houston dataset.}  
    \label{fig:classification_map_Houston}
\end{figure*}

\subsection{Numerical Results}
This subsection discusses our numerical results encompassing comparisons with recently published works. \cref{tab:ip_results} reports numerical results of consolidated algorithms in the IP dataset as the target domain. The efficacy of our algorithm, MIFOMO, is observed where it outperforms other algorithms with a significant gap across all evaluation metrics, i.e., ~$15\%$ margin to MGPDO in the second place in the realm of overall accuracy (OA). This finding clearly demonstrates the efficacy of our mixup domain adaptation strategy combined with a foundation model.  
\cref{tab:pu_results} shows numerical results of the consolidated algorithms in the PU dataset. It is seen that MIFOMO beats MGDPO in the second place with $6\%$ margin. 
\cref{tab:sa_results} exhibits numerical results of the consolidated algorithms, confirming the advantage of our algorithm as the best-performing method with $1.5\%$ gap while 
\cref{tab:hs_results} presents numerical results of the consolidated algorithms in the Houston dataset showing that MIFOMO outperforms CFSL-KT as the second-best method with a notable margin, $13\%$. This finding clearly substantiates our claim that MIFOMO produces state-of-the-art results over prior arts. MIFOMO features a hyper-spectral foundation model with a novel parameter fine-tuning approach, the coalescent projection. It is trained using the mixup domain adaptation method, which distinguishes itself from existing approaches, mostly using the adversarial-based domain adaptation method. That is, no domain discriminator is required in our framework. The mixup domain adaptation approach is underpinned by a two-phase training strategy, i.e., source and intermediate domains, where the intermediate domain functions as a bridge between the source domain and the target domain, enabling seamless knowledge transfer.  

The classification maps of MIFOMO across four datasets are respectively displayed in Fig. \ref{fig:classification_map,IP}, \ref{fig:classification_map,salinas}, \ref{fig:classification_map,pavia}, \ref{fig:classification_map_Houston}, and compared to the ground truth. It is seen that our algorithm is capable of delivering decent classification maps with respect to the ground truth. That is, the classification errors are low, meaning that the predictions are close to the ground truth. 

\begin{table}[!tbp]
    \centering
    \caption{Ablation studies on the Indian Pines dataset.}
    \label{tab:ablation_indian_pines}
    \resizebox{0.5 \textwidth}{!}{
        \begin{tabular}{|l|l|l|l|l|l|l|}
            \toprule
            Label Smoothing & Intermediate Domain & CP & Mixup & OA & AA & KC \\
            \bottomrule
            \cmark & \cmark & \cmark & \cmark & \val{95.44}{3.19} & \val{88.74}{3.93} & \val{94.82}{3.62} \\  \hline
            \xmark & \cmark & \cmark & \cmark & \val{68.22}{3.11} & \val{81.62}{1.89} & \val{64.43}{3.35} \\  \hline
            \cmark & \xmark & \cmark & \cmark & \val{92.29}{4.83} & \val{88.05}{3.92} & \val{91.24}{5.46} \\  \hline
            \cmark & \cmark & \xmark & \cmark & \val{94.57}{5.19} & \val{89.90}{3.75} & \val{93.85}{5.76} \\  \hline
            \cmark & \cmark & \cmark & \xmark & \val{93.25}{4.47} & \val{87.75}{5.72} & \val{90.92}{5.12} \\ \hline
            \xmark & \xmark & \xmark & \xmark & \val{62.48}{3.05} & \val{76.97}{2.02} & \val{58.06}{3.1} \\   \hline
        \end{tabular}
    }
\end{table}

\subsection{Ablation Studies}
Table \ref{tab:ablation_indian_pines} reports numerical results of our ablation study in the Indian Pines dataset, where different configurations of MIFOMO are tested and compared to demonstrate the influence of each learning module of MIFOMO. We commence our ablation study with the label smoothing module, refining the quality of pseudo-labels based on the consistency of neighboring samples. Obviously, the label smoothing approach plays a pivotal role in MIFOMO, where the absence of this module results in a notable performance drop, $17\%$. Note that the pseudo-labels are subsequently applied to create the intermediate domain. Unreliable pseudo labels lead to dramatic performance degradation. The intermediate domain training phase also demonstrates a key contribution to the success of MIFOMO, where its removal is attributed to performance deterioration by about $3\%$. This configuration implies that the model loses the target domain information. The parameter efficient fine-tuning (PEFT) technique via the coalescent projection strategy contributes to $1\%$ performance gain in the IP dataset. This configuration ignores the use of the PEFT technique, where the classification head is the only learnable module during the few-shot learning phase. Last but not least, the use of mixup for the data augmentation method and the establishment of the intermediate domain is linked to $2\%$ performance improvement. The mixup technique allows creations of reliable synthetic samples via linear interpolations of two distinct samples. Note that our mixup strategy is not only limited to the intra-class samples but also the inter-class samples. In other words, the model is trained to recognize new concepts or classes. The use of a foundation model alone doesn't suffice for cross-domain few-shot learning in hyperspectral image classification. Although the foundation model already possesses generalizable features, certain techniques are required to achieve its best potential in downstream tasks.  

\subsection{T-SNE Analysis}
\begin{figure}[htbp]
    \centering
    \subfloat[Indian Pines\label{fig:tsne-ip}]{
        \includegraphics[width=0.45\columnwidth]{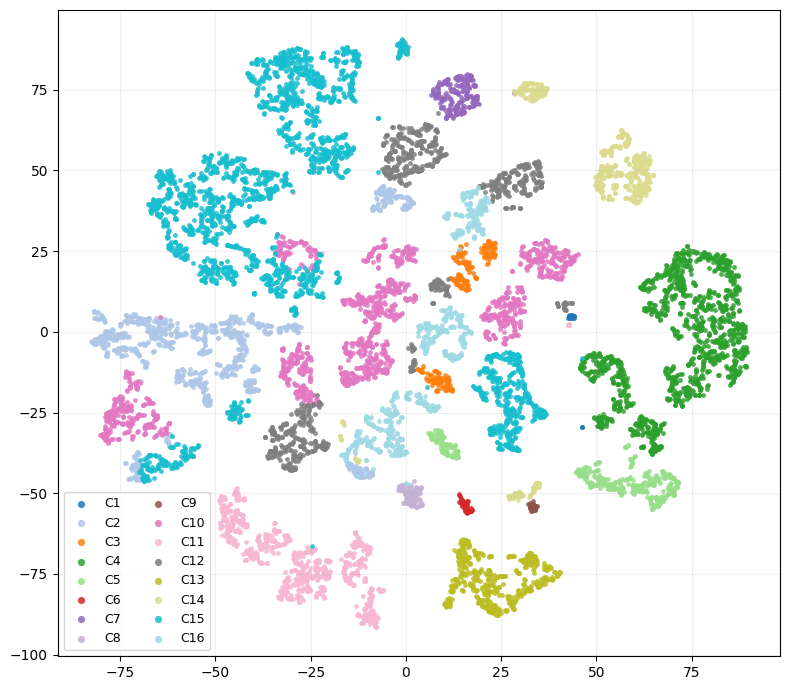}
    }\hfill
    \subfloat[Pavia University\label{fig:tsne-pu}]{
        \includegraphics[width=0.45\columnwidth]{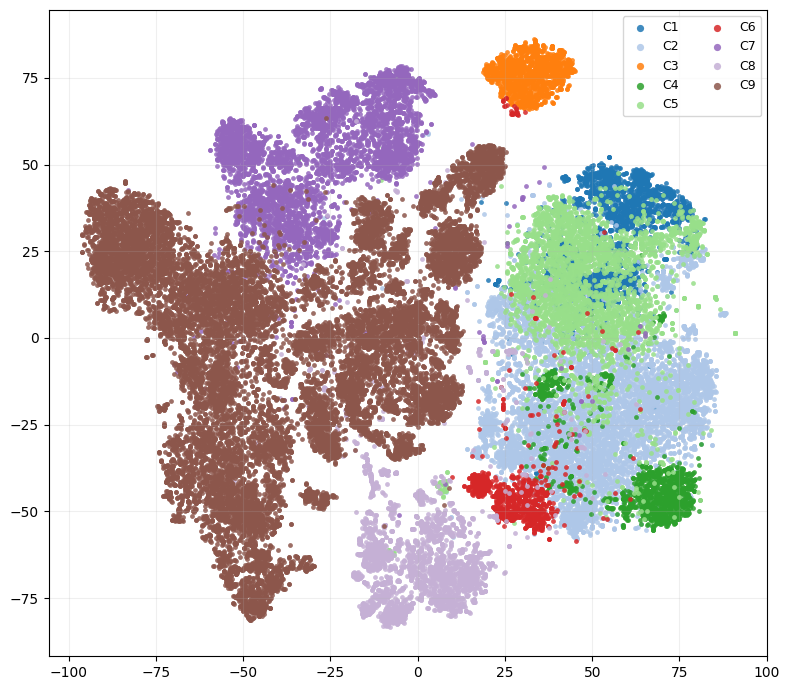}
    }

    \vspace{0.1em}

    \subfloat[Salinas\label{fig:tsne-sal}]{
        \includegraphics[width=0.45\columnwidth]{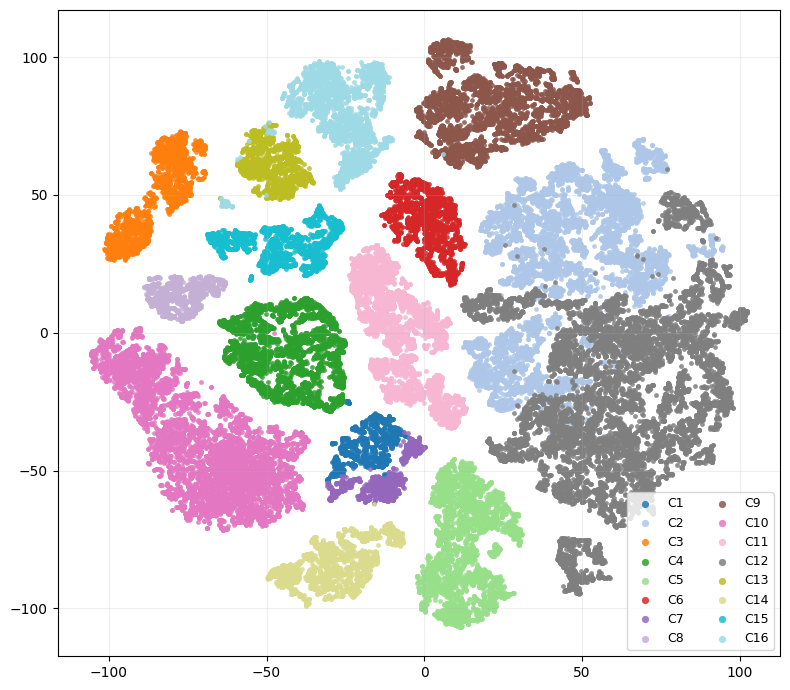}
    }\hfill
    \subfloat[Houston\label{fig:tsne-hou}]{
        \includegraphics[width=0.45\columnwidth]{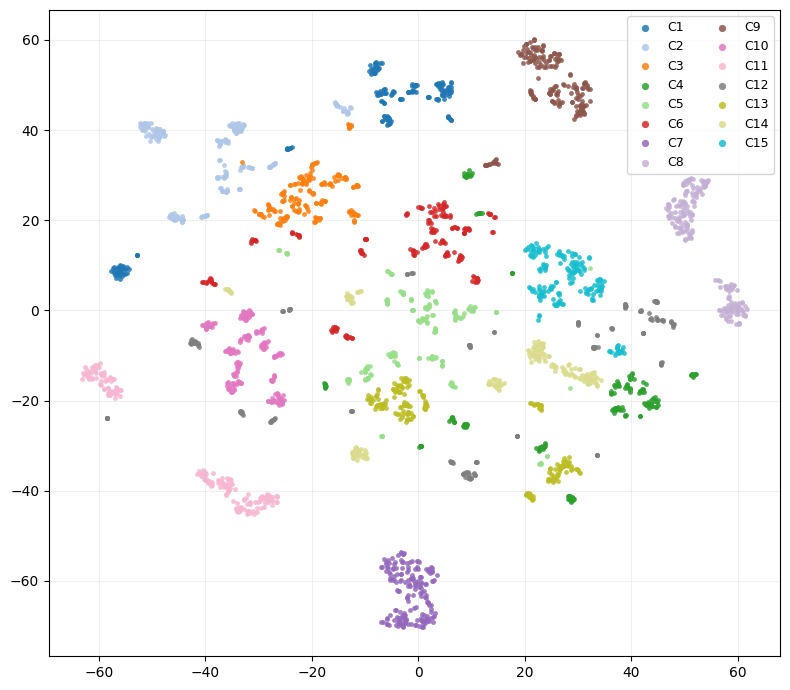}
    }

    \caption{t-SNE plot for the Indian Pines, Pavia University, Salinas, and Houston datasets.}
    \label{fig:t-SNE}
\end{figure}

This section discusses the T-SNE analysis of our algorithm, MIFOMO, and evaluates its embedding quality. Fig.~\ref{fig:t-SNE} exhibits the T-SNE plots across all datasets. It is obvious that MIFOMO generates decent embedding after the training process. That is, different classes can be mapped across different regions in the feature space. In other words, MIFOMO possesses discriminative characteristics promoting correct classification decisions. This finding also demonstrates that the classification error is marginal. 

\section{Conclusion}
This paper addresses the CDFSL problems for HSI classifications via the proposal of the MIxup FOundation MOdel (MIFOMO). MIFOMO is underpinned by a hyper-spectral foundation model, namely HyperSigma, pretrained across a large scale of RS problems via the masked image modeling approach, thus characterizing generalizable features. Our key ideas lie in the concept of coalescent projection (CP) as a parameter-efficient fine-tuning (PEFT) method. To remedy the problem of domain shifts, the mixup domain adaptation method is proposed and introduces an intermediate domain functioning as a bridge between the source and target domains, thus allowing seamless knowledge transfer. The label smoothing method is integrated to avoid noisy pseudo-labels. Our rigorous numerical validations confirm the advantages of our approach, where it outperforms recently published works by up to $14\%$ margins across four benchmark datasets: Indian Pines, Pavia University, Salinas, and Houston. This finding is further confirmed by the ablation study, substantiating positive and cohesive contributions of each learning component. Our T-SNE analysis also shows decent and discriminative embedding qualities of MIFOMO. Albeit promising performances of MIFOMO, it is as with other works in this domain categorized as the transductive approach, where a model has access to unlabeled samples of the target domain, i.e., the query set. Our future work is devoted to studying the inductive method, where a model is disallowed to learn unlabeled samples of the query set, thus transforming it into a much more challenging problem than the transductive setting.

\bibliographystyle{elsarticle-harv}
\bibliography{references}


\end{document}